%% file: main.tex
\documentclass[10pt,twocolumn,letterpaper]{article}

\usepackage[pagenumbers]{cvpr} 

\input{preamble}
\definecolor{cvprblue}{rgb}{0.21,0.49,0.74}
\usepackage[pagebackref,breaklinks,colorlinks,allcolors=cvprblue]{hyperref}
\usepackage{subcaption}
\usepackage{graphicx}
\usepackage{enumitem}
\usepackage{booktabs}
\usepackage{algorithm}
\usepackage{algorithmicx}
\usepackage{algpseudocode}
\usepackage{xcolor} 
\usepackage{colortbl}
\usepackage{multirow}
\usepackage{color}
\usepackage{wrapfig} 
\usepackage{amsthm}
\newtheorem{theorem}{Theorem}

\def\paperID{} 
\def\confName{CVPR}
\def\confYear{2026}

\title{All Vehicles Can Lie: Efficient Adversarial Defense in Fully Untrusted-Vehicle Collaborative Perception via Pseudo-Random Bayesian Inference}

\author{
Yi Yu$^{1}$, Libing Wu$^{1\dagger}$, Zhuangzhuang Zhang$^{1}$, Jing Qiu$^{2}$, Lijuan Huo$^{1}$, Jiaqi Feng$^{1}$ \\
$^{1}$School of Cyber Science and Engineering, Wuhan University\\
$^{2}$Cyberspace Institute of Advanced Technology, Guangzhou University\\
{\tt\small \{yui1212, wu, zhzhuangzhuang, lijuanHuo, jiaqiFeng\}@whu.edu.cn}, {\tt\small \ qiujing@gzhu.edu.cn}
}

\begin{document}
\maketitle
\renewcommand{\thefootnote}{}
\footnotetext{\textdagger\ Corresponding author.}
\renewcommand{\thefootnote}{\arabic{footnote}}
\input{sec/0_abstract}    
\input{sec/1_intro}
\input{sec/2_relatedwork}
\input{sec/3_attackformulation}
\input{sec/4_method}
\input{sec/5_theory}
\input{sec/6_experiments}
\input{sec/7_conclusion}

{
    \small
    \bibliographystyle{ieeenat_fullname}
    \bibliography{main}
}

\input{sec/X_suppl}

\end{document}

%% file: sec/0_abstract.tex
\begin{abstract}
Collaborative perception (CP) enables multiple vehicles to augment their individual perception capacities through the exchange of feature-level sensory data. However, this fusion mechanism is inherently vulnerable to adversarial attacks, especially in fully untrusted-vehicle environments. Existing defense approaches often assume a trusted ego vehicle as a reference or incorporate additional binary classifiers. These assumptions limit their practicality in real-world deployments due to the questionable trustworthiness of ego vehicles, the requirement for real-time detection, and the need for generalizability across diverse scenarios. To address these challenges, we propose a novel Pseudo-Random Bayesian Inference (PRBI) framework, a first efficient defense method tailored for fully untrusted-vehicle CP. PRBI detects adversarial behavior by leveraging temporal perceptual discrepancies, using the reliable perception from the preceding frame as a dynamic reference. Additionally, it employs a pseudo-random grouping strategy that requires only two verifications per frame, while applying Bayesian inference to estimate both the number and identities of malicious vehicles. Theoretical analysis has proven the convergence and stability of the proposed PRBI framework. Extensive experiments show that PRBI requires only 2.5 verifications per frame on average, outperforming existing methods significantly, and restores detection precision to between 79.4\% and 86.9\% of pre-attack levels.
\end{abstract}

%% file: sec/1_intro.tex
\section{Introduction}
\label{sec:intro}

Multi-vehicle collaborative perception (CP) enables connected and autonomous vehicles (CAVs) to extend their perceptual range by sharing sensor data via vehicle-to-vehicle (V2V) communication~\cite{gao2024survey, fang2025prioritized}. This mitigates blind spots inherent to single-vehicle sensing. Generally, CP can be implemented through raw data-level (early), feature-level (intermediate), or decision-level (late) fusion, with intermediate fusion offering the best trade-off between efficiency and accuracy~\cite{liu2024survey, hu2024communication, wang2023umc}. However, it also exposes the ego vehicle to increased security risks. Malicious collaborators can inject adversarial perturbations into shared feature maps, causing substantial perception errors when fused by the ego vehicle~\cite{tu2021adversarial}. Effective defense thus hinges on detecting and isolating such malicious vehicles.

Sampling-based defenses~\cite{hu2025cp, li2023among} have been proposed to mitigate adversarial threats in CP.
These methods typically validate information consistency between the ego vehicle and sampled collaborators via iterative consensus algorithms. Although demonstrating robustness against specific attack patterns, their effectiveness fundamentally depends on two strong assumptions: (1) the absolute reliability of ego vehicle inputs, and (2) prior knowledge of attacker conditions. Such requirements significantly constrain their operational adaptability and computational efficiency.
In addition, classifier-based approaches~\cite{zhao2024made, hu2025cp+, tao2025gcp} train binary classifiers to distinguish malicious collaborators from benign ones based on statistical patterns such as feature residual distributions between benign (ego) and malicious vehicles. While offering greater flexibility, these methods expand the attack surface, struggle to generalize to unseen scenarios, and impose significant computational overhead, thereby posing substantial challenges to practical deployment. Accordingly, this paper focuses on sampling-based validation as a more robust and lightweight alternative. 

However, a fundamental limitation common to all existing methods is the assumption that the ego vehicle is inherently trustworthy and its perception results can serve as a reliable reference. In reality, non-ego vehicles can be directly manipulated by attackers, who inject adversarial perturbations into their feature maps and subsequently transmit the tampered data to the ego vehicle. For the ego vehicle, although attackers cannot directly compromise its internal system, they can still disrupt its feature map via LiDAR injection attacks~\cite{jin2023pla, sato2025realism} or data interception attacks~\cite{cao2023you}. Thus, this assumption is misaligned with real-world scenarios, where the CP environment is a fully untrusted-vehicle setting and all vehicles (including the ego one) must be treated as potentially malicious entities capable of transmitting adversarial features~\cite{wang2024break, zhang2024perception, lou2024first}. In short, \textit{\textbf{All Vehicles Can Lie}} in realistic scenarios. This significantly complicates the task of detecting malicious feature maps within the ego vehicle. Moreover, the verification count per frame for both sampling- and classifier-based approaches increases linearly with the total number of vehicles, leading to a low detection efficiency. In summary, achieving \textbf{robust adversarial defense} with \textbf{minimal verification cost} remains a critical challenge in CP, where \textit{all vehicles can lie}.

To tackle this, it is essential to first establish a realistic and robust reference signal. Inspired by the inherent spatial continuity between adjacent frames~\cite{li2024simdiff, akhtar2024inter}, we hypothesize that LiDAR-based perception outputs across successive frames also exhibit a consistent degree of similarity that exceeds a critical threshold. To verify this hypothesis, we compute the Jaccard similarity~\cite{niwattanakul2013using} between perception outputs across consecutive frames under various normal and adversarial settings (see appendix for experimental details). Extensive evaluations show that inter-frame similarity is stably around $0.8$ in benign scenarios, whereas adversarial settings cause it to drop sharply. This pronounced and consistent gap confirms that temporal consistency can serve as a reliable self-referential signal even without assuming any trusted vehicle, enabling us to replace the ego-trust assumption with a frame-wise self-supervision mechanism that uses the prior frame as a probabilistic reference.

Building on the validated hypothesis, we propose the novel \textbf{P}seudo-\textbf{R}andom \textbf{B}ayesian \textbf{I}nference (PRBI) defense framework, a mathematically principled method for efficient and assumption-free detection of malicious vehicles in fully untrusted-vehicle CP without prior knowledge of attackers. 
Specifically, we design a \textbf{pseudo-random grouping strategy} that requires only two additional verifications per frame, completely decoupling the number of verifications from the total number of vehicles and significantly reducing detection costs. Additionally, we mathematically model this grouping strategy as a repeated random sampling process, estimating the number of malicious vehicles through mathematical approximation and evaluating the benign probability of each vehicle using \textbf{Bayesian probabilistic inference}. By combining the estimated number of malicious vehicles with the benign probability of individual cars, suspected malicious vehicles can be identified and filtered out. To further enhance detection efficiency, we incorporate \textbf{hypothesis testing} techniques, enabling PRBI to preemptively assess convergence and prioritize the fusion of features from vehicles deemed trustworthy during the non-convergence stage. Meanwhile, we rigorously validate the PRBI framework through formal mathematical proofs of its convergence and rounding operation properties, thereby establishing its theoretical guarantees for attacker detection.

Our main contributions are as follows:
\begin{itemize}[leftmargin=*, topsep=0pt]
\item We analyze inter-frame perceptual similarity and \textbf{first} exploit it as a self-referential defense signal for \textbf{fully untrusted-vehicle} CP, which addresses the core challenge when \textit{all vehicles can lie}.

\item We propose the \textbf{Pseudo-Random Bayesian Inference (PRBI)} framework, achieving robust fusion and attacker detection without prior knowledge, averaging only two verifications per frame.

\item We theoretically prove PRBI's convergence and validity from \textbf{a mathematical perspective} and conduct a comprehensive evaluation to demonstrate its effectiveness. 
\end{itemize}

%% file: sec/2_relatedwork.tex
\section{Related Work}

\textbf{Adversarial CP}. Recent studies reveal that CP is more susceptible to adversarial attacks than single-agent systems~\cite{tsai2020robust, schiegg2020analytical, wang2025imperceptible, yu2025enduring, wang2025threat}. In feature-level fusion, attackers can embed malicious perturbations into shared feature maps via white-box methods (e.g., PGD, C\&W)~\cite{tu2021adversarial, zhang2024data}, severely degrading perception accuracy. Existing defenses such as adversarial training~\cite{madry2017towards} suffer from limited generalization and high computational costs, while detection-based approaches~\cite{zhao2024made, hu2025cp+, tao2025gcp} struggle against adaptive attacks.

\textbf{Defensive CP}. Defense strategies can be broadly categorized into sampling-based and classifier-based approaches.  
Sampling-based methods~\cite{hu2025cp, li2023among, zhang2024data} enforce consistency between the ego vehicle and a subset of collaborators via iterative consensus protocols. While offering partial robustness, they rely on strong trust assumptions and adversarial priors, limiting applicability in fully untrusted environments.  
Classifier-based defenses~\cite{zhao2024made, hu2025cp+, tao2025gcp} detect malicious participants by learning discriminative patterns in residuals between shared and local features; however, their performance degrades when multiple vehicles are compromised or under diverse driving conditions. 
Recent work on security-aware sensor fusion further explores probabilistic trust modeling. 
MATE~\cite{hallyburton2025security} proposes a geometry-based multi-agent trust estimator that updates trust scores according to observation consistency, requiring object tracking and visibility reasoning. 
In contrast, PRBI is geometry-free and leverages self-referential temporal consistency across pseudo-random collaborator subsets, enabling attack detection without explicit scene modeling, tracking, or ego-trust assumptions. 
By removing prior trust requirements and attacker assumptions, PRBI provides mathematically grounded robustness with constant detection overhead.

%% file: sec/3_attackformulation.tex
\section{Attack Formulation}
\subsection{Attack Model}

\begin{figure*}[!t]
\centering
\includegraphics[width=0.75\textwidth]{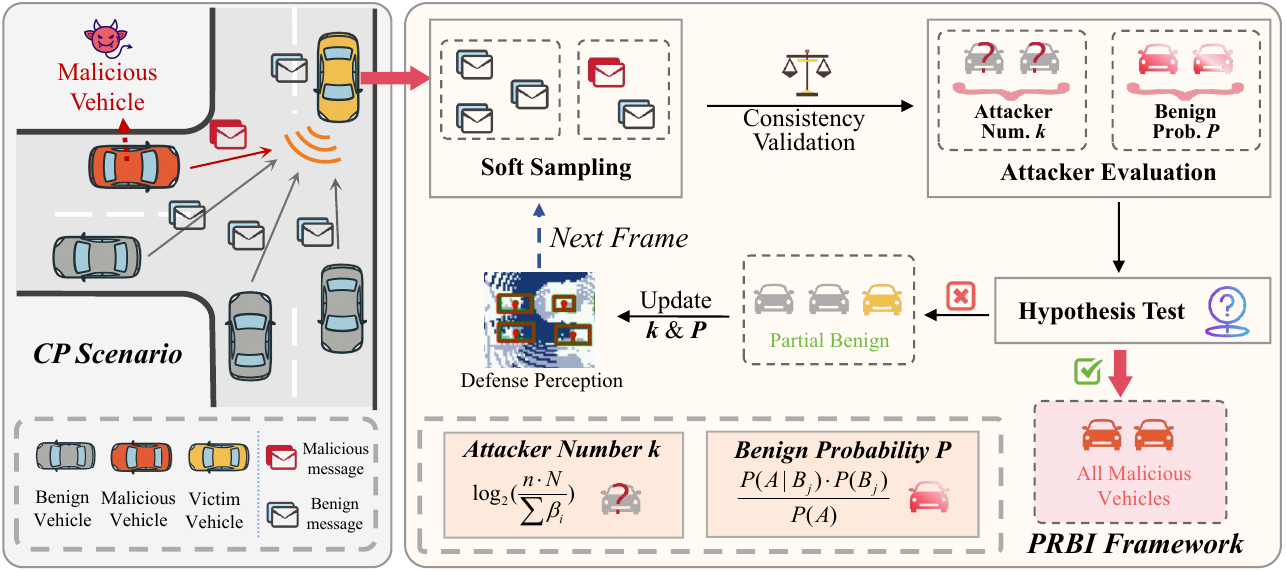}
\vspace{-2mm}
\caption{The Pseudo-Random Bayesian Inference (PRBI) framework methodology overview.}
\vspace{-0.5em}
\label{fig:framework}
\end{figure*}

We focus on a vehicle-to-vehicle (V2V) collaborative perception (CP) scenario, as shown in~\cref{fig:framework}. Similar to previous defensive work, we assume that all vehicles employ the same LiDAR-based feature-level CP model, and any feature map may be perturbed.  
Before an attack, the attacker selects a fixed set of malicious vehicles (including the ego vehicle), constrained by the limited budget to compromise all participants. During an attack frame, all malicious vehicles are controlled to produce adversarial features. Perception errors arise when the ego fuses these corrupted features with its own. The defense objective
is thus to detect and exclude malicious feature contributions before fusion.

\subsection{Problem Formulation}
We consider a collaborative scenario involving $n$ CAVs, denoted as $a_1, a_2, \dots, a_n$, where $a_1$ represents the ego vehicle and $n \geq 2$. Each vehicle is equipped with the same feature encoder $f_\mathcal{E}(\cdot)$, feature fusion module $f_\mathcal{F}(\cdot)$, and object detection decoder $f_\mathcal{D}(\cdot)$. At time frame $t$, vehicle $a_i$ captures LiDAR point cloud raw data $C_i^t$, which is encoded into a feature map: $F_i^t = f_\mathcal{E}(C_i^t) \in \mathbb{R}^{W \times H \times C}$, where $W$, $H$, and $C$ denote the width, height, and number of channels of the feature map, respectively. 
Vehicles broadcast their encoded features to collaborators. Upon reception, $a_i$ transforms received maps into its local coordinate frame via $f_\mathcal{T}(\cdot)$ and fuses them with its own:  
$
H_i^t = f_\mathcal{F}\big(F_i^t \cup f_\mathcal{T}(\{F_j^t\}_{j \neq i})\big).
$
The fused representation is then decoded to yield perception outputs  
$
D_i^t = f_\mathcal{D}(H_i^t) = \{d_j\}_{j\in[1,L]},
$
where $L$ is the number of predicted bounding boxes, each $d_j \in \mathbb{R}^{c+l}$ containing $c$ class logits and $l$ location parameters.

\textbf{Adversarial Perception}. In a fully untrusted-vehicle setting, the victim’s own feature map may also be compromised without its awareness. The adversary arbitrarily compromises $k$ out of $n$ vehicles, $1 \leq k \leq n-1$, forming the malicious set $M$. 
The attack aims to mislead the ego’s perception by injecting bounded perturbations $\delta \in \mathbb{R}^{W \times H \times C}$ into the feature maps $\{F_i^t\}_{i \in M}$, with the dual objectives of concealing real targets and inducing false object perceptions.
After fusion, the ego’s perception output becomes:
\begin{equation}
D_1^{t}(\delta) = f_{\mathcal{D}} \circ f_{\mathcal{F}} \circ f_{\mathcal{T}}\big(\{F_i^t + \delta\}_{i \in M} \cup \{F_j^t\}_{j \notin M}\big) = \{d_j^{\prime}\}_{j \in [1,L]}.
\label{eq:output}
\end{equation}
The perturbation is bounded by $\|\delta\|\leq \Delta$ and optimized to maximize divergence as follows:
\begin{equation}
\max_{\|\delta\|\leq \Delta} \sum_{j=1}^{L} \mathcal{L}_\mathrm{det}(d_j, d_j^{\prime}),
\label{eq:advloss}
\end{equation}
where $\mathcal{L}_\mathrm{det}$ is the multi-agent standard detection loss following~\cite{zhao2024made,tu2021adversarial} and $\Delta$ constrains perturbation magnitude.

%% file: sec/4_method.tex
\section{PRBI Defense Method}
\subsection{Defense Challenges and Key Findings}
\label{sec:defense_challenges}
\label{sec:findings}

In fully untrusted-vehicle CP scenarios, designing an effective defense faces three key challenges:
\textbf{C1: Detection without trust assumptions.}
The defender has no prior knowledge of which vehicles are malicious, their number or proportion, and even the ego vehicle may be compromised, making it difficult to detect attackers without any trust assumption.
\textbf{C2: Scalability of per-frame defense cost.}
Existing sampling or verification methods incur per-frame costs that grow linearly with fleet size, which is impractical for real-time large-scale CP. Achieving constant-cost detection per frame is essential.
\textbf{C3: Fast and complete attacker identification.}
Randomized sampling may delay attacker exclusion, causing prolonged risk exposure. The third key challenge is to guarantee both high detection speed and completeness despite sampling randomness.

\textit{To tackle these challenges, we derive two key findings:}
\textbf{F1: Inter-frame similarity divergence enables ego-agnostic detection.} To address \textbf{C1}, we conduct a dedicated experiment analyzing the similarity of perception results between consecutive frames and obtain the following key observation: benign perception exhibits significantly higher inter-frame similarity compared to adversarial perception (see the appendix for details). We conclude that benign perception maintains an average similarity of around 0.8, while adversarial cases rarely exceed 0.3. This property allows using the previous frame’s verified benign output as a trusted reference for current-frame detection, removing dependency on prior trust assumptions.
\textbf{F2: Binary grouping approximates random sampling and implicitly encodes key information about malicious vehicles.} Theoretically, the minimum cost for additional verification per frame is two verifications. 
\begin{table}\small 
    \centering 
    \caption{Comparison between the theoretical probability of sampling an all-benign group and the empirical probability obtained by randomly splitting vehicles into two groups. The frame count is recorded once the observed probability $\eta$ converges to $P_{ideal}^\prime$.}
    \vspace{-2.5mm}
    \label{tab:probapp}
    \begin{tabular}{@{}c|cccc@{}}
        \toprule
        \textbf{Attacker Number ($n=5$)} & \textbf{1} & \textbf{2} & \textbf{3} & \textbf{4} \\ 
        \midrule
        Theoretical Prob. $P_{ideal}$ & 0.484 & 0.226 & 0.097 & 0.032 \\
        Approximate Prob. $P_{ideal}^\prime$ & 0.500 & 0.250 & 0.125 & 0.063  \\
        \midrule
        Validation Frame Count & 40 & 29 & 28 & 21\\
        Practical All-Benign Prob. $\eta$ & 0.491 & 0.258 & 0.103 & 0.045\\
        \bottomrule
    \end{tabular}
    \vspace{-3mm}
\end{table}
To approach this optimal cost, we divide all vehicles into two random groups per frame and validate each group's collaborative perception result against the benign perception of the previous frame. This yields two key insights: \textit{(1) Binary grouping statistically resembles random sampling without replacement}, where the probability of obtaining an all-benign group is:
\begin{equation}
    P_{ideal} = \frac{2^{n-k}-1}{2^n-1} \approx P_{ideal}^\prime = \frac{2^{n-k}}{2^n}.
    \label{pideal}
\end{equation}
The experiments shown in~\cref{tab:probapp} confirm that in finite frames, binary grouping closely approximates this process ($\eta \approx P_{ideal}^\prime$), with $P_{ideal}^\prime$ providing a tractable estimate dependent only on the attacker number $k$. 
\textit{(2) Binary group verification reveals per-vehicle trustworthiness}.
Each group verification yields not only a group-level result but also a single validation outcome per participating vehicle.
By maintaining two counters per vehicle (one for successful benign and one for failed malicious validations),
we track every vehicle’s verification history. The global empirical benign ratio $\eta$ of normal validations can then
be computed using the verification history and approximated to $P_{ideal}^\prime$. From this, we can infer the number of attackers as follows:
\begin{equation}
    \label{eq:kkk}
    k = \log_2(\frac{1}{P_{ideal}^\prime}) \approx \log_2(\frac{1}{\eta}).
\end{equation}
Furthermore, a vehicle involved in almost no benign verifications is likely to be malicious. Thus, by combining the estimated $k$ and per-vehicle validation statistics, we can identify likely attackers.
In conclusion, binary grouping not only addresses the core of \textbf{C2} but also enables efficient, cost-effective defense which is essential for \textbf{C3}.

\vspace{-1mm}
\subsection{PRBI Overview}
\label{sec:over}
As observed in \textbf{F1}, the distribution of perception bounding box similarity between adjacent frames differs significantly under normal versus adversarial conditions. We quantify this using Jaccard similarity, defined for two consecutive detection sets $D^t$ and $D^{t-1}$ as:
\begin{equation}
    \operatorname{Jaccard}(D^t, D^{t-1}) =
    \frac{|\mathcal{M}|}{|D^{t}| + |D^{t-1}| - |\mathcal{M}|},
    \label{eq:jaccard}
\end{equation}
where $\mathcal{M}$ is the set of matched pairs (Hungarian matching with IoU above a threshold~\cite{keswani2022proto2proto}) and $|\mathcal{M}|$ denotes the number of valid matches.
Leveraging this distributional gap, we assume the initial frame's perception is entirely correct. For subsequent frames, if the similarity with the previous frame falls below a warning threshold $\epsilon$, the frame is flagged as potentially attacked. Then, building upon \textbf{F2}, PRBI detects and excludes malicious vehicles through following steps:
\begin{itemize}[leftmargin=*, topsep=0pt]
    \item {\textbf{Step 1: Initialization.} At $t=0$, initialize two $n$-dimensional arrays: normal detection counts $\mathbf{c}_{\text{normal}} = (\beta_1,\dots,\beta_n)$ and abnormal detection counts $\mathbf{c}_{\text{abnormal}} = (N-\beta_1,\dots,N-\beta_n)$, with $N$ (total detections) and $\beta_i$ set to 0, $n$ represents the total number of vehicles.}

    \item \textbf{Step 2: Soft Sampling \& Consistency Validation.} For each frame, the victim vehicle partitions received shared features into two groups using \textbf{soft sampling} strategy, performs perception fusion, and computes similarity with the previous benign frame. Groups exceeding $\epsilon$ have their members' $\beta_i$ incremented. Then $N$ and both count arrays ($\mathbf{c}_{\text{normal}}$, $\mathbf{c}_{\text{abnormal}}$) are updated.

    \item {\textbf{Step 3: Attacker Evaluation.} Using $\mathbf{c}_{\text{normal}}$ and $\mathbf{c}_{\text{abnormal}}$, apply probabilistic approximation and Bayesian probabilistic inference to estimate the number of suspicious malicious vehicles $m$ and each vehicle’s benign probability $P_\text{benign}=\left(p_1,...,p_n\right)$. Meanwhile, select the $m$ vehicles with the lowest benign probabilities as the estimated malicious vehicles, thus addressing \textbf{C1} and \textbf{C2}.}

    \item \textbf{Step 4: Hypothesis Testing \& Defense Perception.} A T-test is applied to determine if estimates have converged (addressing \textbf{C3}). If converged, the attacker set $M=\{\text{argsort}(P_{\text {benign}})_{0:\left\lfloor m \rceil\right.}\}$ and count $m$ are output; otherwise, vehicles with $p_i \neq 0$ serve as collaborators in the next frame and return to step 2. Since $\beta_i=0$ for any true attacker, its $p_i$ remains zero, ensuring eventual exclusion. More details are provided in~\cref{sec:prbi}.
\end{itemize}

The convergence of $m$ to the true number of attackers $k$ is critical to the effectiveness of PRBI, and the corresponding theorems are provided in~\cref{sec:proof}.

\vspace{-1mm}
\subsection{PRBI Framework Design}
\label{sec:prbi}
Given reference perception $D_{\text{ref}}$ from the previous benign frame, PRBI executes four core steps: (i) soft sampling via pseudo-random grouping, (ii) group consistency validation, (iii) attacker estimation, and (iv) convergence testing. The pseudocode of PRBI is provided in the appendix.

\textbf{Soft Sampling.} To minimize verification costs, we propose a pseudo-random grouping based on \textbf{F2}. This strategy divides all shared information into two groups and performs only two additional verifications per frame. Specifically, at each frame, we divide the most suspicious $\left\lfloor m \right\rfloor$ vehicles into one group and the rest into the second group to approximate the random sampling process: 
\begin{equation}
    (Group_1,Group_2)=(\pi_{0:\lfloor m \rfloor},\pi_{\lfloor m \rfloor : n}), 
\end{equation}
where $\pi=\text{argsort}(P_{\text{benign}})$ and $m$ is the current estimated number of attackers.
This method has two main advantages: First, by simulating the sampling-without-replacement process, it can effectively uncover hidden information (see \textbf{F2}). Second, the pseudo-random grouping strategy based on $m$ and $P_{\text{benign}}$ ensures that the final estimated number of attackers $m$ converges to the actual number $k$. Additionally, to guarantee convergence, $m$ must be floored, and other rounding operations may cause convergence bias. A detailed proof of these conclusions is provided in~\cref{sec:proof}.

\textbf{Consistency Validation.} Based on the previous frame’s reliable perception results $D_\text{ref}$, the two groups are sequentially subjected to consistency validation. For each group, feature-level collaborative perception is first performed, and its consistency with the reference result is then computed as $\operatorname{Jaccard}(D_{\text{Group}_{}}, D_{\text{ref}})$. If the consistency value exceeds the threshold $\epsilon$, it indicates that the group likely contains no malicious vehicles, and the normal detection count for each vehicle in the group $\{\beta_i|i\in\text{Group}\}$ is incremented by 1. After both groups are validated, the total number of detections $N$ for each vehicle is increased by 1, and the arrays $\mathbf{c}_{\text{normal}}$ and $\mathbf{c}_{\text{abnormal}}$ are updated in sequence. In other words, the consistency validation results are used to maintain the updates of $\mathbf{c}_{\text{normal}}$ and $\mathbf{c}_{\text{abnormal}}$, which are critical for the subsequent evaluation of potential attackers. 

\textbf{Attacker Evaluation.} We first estimate the number of attackers. Assuming that in each round, a group of vehicles is randomly sampled from all vehicles, and based on the idealized random process (see~\cref{pideal}), the probability $\rho$ of sampling an all-benign group can be expressed as:
\begin{equation}
    \rho=\frac{2^{n-k}-1}{2^{n}-1} \approx \frac{2^{n-k}}{2^{n}}=2^{-k},
\end{equation}
where $n$ denotes the total number of vehicles and $k$ is the true number of attackers. Meanwhile, using $\mathbf{c}_{\text{normal}}$ and $\mathbf{c}_{\text{abnormal}}$, the empirical ratio $\eta$ of normal detection counts can be calculated as:
\begin{equation}
    \eta=\frac{\text{Benign\ Counts}}{\text{Total\ Counts}}=\frac{\sum \mathbf{c}_{\text{normal}}}{{\sum_{i=1}^n}N}=\frac{\sum_{i=1}^{n}\beta_i}{n \cdot N}.
\end{equation}
Through \textbf{F2} in~\cref{sec:findings}, we can approximate $\eta$ to $\rho$ and estimate the number of attackers like~\cref{eq:kkk}. Thus, we assume $\eta \approx \rho$, and the number of attackers can be estimated as:
\begin{equation}
    m = \log_2 \left( \frac{1}{\rho} \right) \approx \log_2 \left( \frac{1}{\eta} \right) = \log_2 \left( \frac{n \cdot N}{\sum_{i=1}^n \beta_i} \right).
    \label{eq:m_compute}
\end{equation}
After estimating the number of attackers, the next step is to identify specific malicious vehicles by computing the \textit{benign probability} $P_{\text{benign}}[j]$ of each vehicle based on Bayes’ theorem. Let $\mathcal{A}$ denote the event that an attack occurs and $\mathcal{B}_j$ the event that vehicle $j$ is benign. According to Bayes’ rule, the posterior benign probability is given by:
\begin{equation}
\label{eq:bayes}
P_{\text{benign}}[j] =P(\mathcal{B}_j|\mathcal{A})= \frac{P(\mathcal{A}|\mathcal{B}_j) \cdot P(\mathcal{B}_j)}{P(\mathcal{A})}.
\end{equation}
Here, $P(\mathcal{A}|\mathcal{B}_j)$ represents the likelihood of observing an attack given that vehicle $j$ is benign, which is estimated as:
\begin{equation}
\label{eq:a_given_b}
P(\mathcal{A}|\mathcal{B}_j) = \frac{\sum \mathbf{c}_{\text {abnormal}}-\mathbf{c}_{\text{ abnormal}}[j]}{\sum_{i=1}^n N-N}.
\end{equation}
The prior benign probability $P(\mathcal{B}_j)$ is defined as a weighted sum of the short-term benign probability from the previous frame and the long-term normal detection ratio:
\begin{equation}
\label{eq:b_prob}
P(\mathcal{B}_j) = \gamma\cdot P_{\text {benign}}[j]+\lambda\cdot\frac{\mathbf{c}_{\text {normal}}[j]}{N},
\end{equation}
where $P_{\text{benign}}$ is initialized to \textbf{0} at first frame, and $\gamma$ and $\lambda$ are the respective weights for short-term and long-term components. This design mitigates the influence of early unstable groupings by incorporating temporal memory. Finally, the global probability of an attack $P(\mathcal{A})$ is given by the proportion of abnormal detections in the entire system $P(\mathcal{A}) = \frac{\sum \mathbf{c}_{\text {abnormal}}}{\sum_{i=1}^n N}$. Based on the estimated number of attackers $m$, the set of suspected malicious vehicles consists of the $\left\lfloor m \right\rceil$ vehicles with the lowest benign probabilities:
\begin{equation}
    Attackers=\text{argsort}(P_{\text{benign}})_{0:\left\lfloor m \right\rceil}.
    \label{eq:attackers}
\end{equation}

\textbf{Hypothesis Testing.} By iteratively executing the above steps, the estimated attacker number $m$ eventually converges to the true attacker count $k$, and the set $Attacker$ is guaranteed to include all malicious vehicles (see mathematical theorem in~\cref{sec:proof}).
To eliminate redundant group verification, PRBI determines whether $m$ has converged using the hypothesis testing method in advance.
The reason for using hypothesis testing is that we observe that $m$ fluctuates within a small range around $k$, and when this fluctuation remains confined within a certain interval for many frames, it can be considered as convergence. Hypothesis testing is well-suited for this. Specifically, we maintain a queue $W$ of window size $w$, storing the estimated $m$ values from the past $w_p$ frames as sample data ($w_p \le w$, $w_p$ is as large as possible).
Then, in the current frame, we define the null hypothesis $H_0:k=m$ and the alternative hypothesis $H_1:k \neq m$. The sample mean is $\bar{x}=\frac{1}{w_p}\sum_{i=1}^{w_p} W_i$, and the sample variance is $S^2=\frac{1}{w_p-1}\sum_{i=1}^{w_p}\left(W_i-\bar{x} \right)^2$. Next, we perform a T-test with confidence level $\alpha$ to compute the rejection region $\mathcal{W}$ and test $H_0$:
\begin{equation}
    \mathcal{W}=\{|\frac{\bar{x}-m}{S/\sqrt{w_p}}|>t_{\frac{\alpha}{2}}(w_p-1)\},
\end{equation}
where $t_{\frac{\alpha}{2}}(w_p-1)$ denotes the upper quantile of the t-distribution with $w_p-1$ degrees of freedom. To prevent the slow convergence of $m$ from falsely causing the T-test to accept $H_0$, we introduce a second convergence condition. For a malicious vehicle $j$, the number of normal detections $\beta_j$ must be 0. Hence, both $P(\mathcal{B}_j)$ and $P_{\text{benign}}[j]$ must also be 0 (as derived from~\cref{eq:b_prob} and~\cref{eq:bayes}). When $m$ converges, its value must equal the number of vehicles with zero benign probability in $P_\text{benign}$. This forms the second convergence condition. When both conditions are satisfied, the current $Attacker$ set is exactly the set of all malicious vehicles, indicating a successful detection.
If the convergence conditions are not met, then we select vehicles with non-zero benign probability as collaborators for this frame to perform perception (according to the second convergence condition), and use their benign perception results to update the reference for the next frame’s detection process.

When abnormal frames first appear, both sampled groups may contain malicious vehicles, resulting in all vehicles having zero benign probability. To select valid collaborators in this frame, the divide-and-conquer method is used to recursively search for the first valid subset of benign vehicles in the two groups to serve as collaborators for the current frame, while also updating $\mathbf{c}_{\text{normal}}$, $N$, and $\mathbf{c}_{\text{abnormal}}$.


%% file: sec/5_theory.tex
\section{Theoretical Results}
\label{sec:proof}
In this section, we present our main theoretical conclusions about the convergence of the estimated attacker number $m$ under the proposed \textbf{pseudo-random grouping} strategy, as well as the effect of different rounding schemes. The complete proofs are provided in the appendix.
\begin{theorem}[Convergence of $m$ to $k$]

\label{thm:convergence}
Under the pseudo-random grouping strategy, the estimated number of attackers $m$ will monotonically converge to the true number of attackers $k$. 
\end{theorem}
\begin{theorem}[Effect of Rounding Strategy]

\label{thm:rounding}
Let $\lfloor \cdot \rfloor$, $\lceil \cdot \rceil$, and $\lfloor \cdot \rceil$ denote floor, ceiling, and nearest-integer rounding.  
Only the floor rounding $\lfloor m \rfloor$ guarantees the exact convergence of $m$ to $k$ under pseudo-random grouping.  
The nearest-integer and ceiling rounding result in convergence to $k - 0.5$ and $k - 1$, respectively. Note that when the true number of attackers $k = 1$ with ceiling rounding, $m$ will not tend toward 0. In this case, the final $m$ will converge to $\log_2(\frac{n}{n-1})$.
\end{theorem}

%% file: sec/6_experiments.tex
\section{Experiments}

\subsection{Experimental Setups}
We evaluate PRBI on the V2X-Sim dataset~\cite{li2022v2x}, which provides multi-vehicle LiDAR point clouds in the nuScenes format~\cite{caesar2020nuscenes}. Detection performance is measured by Average Precision (AP) at IoU thresholds of 0.5 and 0.7, while efficiency is quantified by the number of additional verification comparisons per frame.
For adversarial settings, we adopt BIM~\cite{kurakin2018adversarial}, C\&W~\cite{carlini2017towards}, and PGD~\cite{madry2017towards}. The collaborative perception framework is based on V2VNet~\cite{wang2020v2vnet} for feature fusion and FaFNet~\cite{luo2018fast} as the detection backbone. To assess the robustness of PRBI under different fusion paradigms, we also compare simple fusion strategies (mean, max, and sum) with the advanced DiscoNet~\cite{li2021learning}. The single-vehicle lower-bound baseline follows the feature mean fusion setting~\cite{xiang2023multi}.
We further compare PRBI with the SOTA defenses: ROBOSAC~\cite{li2023among} and PASAC~\cite{hu2025cp}.

\subsection{Defense Effectiveness}

\begin{table}\fontsize{8.5pt}{9.5pt}\selectfont
    \caption{Comparison of the verification count per frame. The best results are marked in \textcolor{cyan!60!black}{\text{blue}}. PRBI yields the lowest detection cost.}
    \label{tab:vercount}
    \centering
    \vspace{-3mm}
    \setlength{\tabcolsep}{4.5pt}
    \begin{tabular}{@{}l|{c}|cccc|c@{}}
        \toprule
        \multirow{3}{*}{\textbf{Methods}} & \multirow{3}{*}{\textbf{Metric $\downarrow$}} 
        & \multicolumn{4}{c|}{\textbf{Attacker Ratio (\%)}} 
        & \multirow{3}{*}{\textbf{Avg. $\downarrow$}} \\
        \cmidrule(lr){3-6}
        & & 20 & 40 & 60 & 80 &  \\
        \midrule
        \multirow{3}{*}{ROBOSAC~\cite{li2023among}}    
        & Min & 1   & 1   & 1   & 1   & 1.0 \\
        & Max & 19  & 39  & 46  & 17  & 30.3 \\
        & Avg & 4.89 & 10.36 & 8.29 & 4.73 & 7.1 \\
        \midrule
        
        \multirow{3}{*}{PASAC~\cite{hu2025cp}}   
        & Min & 4   & 4   & 6   & 8   & 5.5 \\
        & Max & 6  & 8  & 8  & 8  & 7.5 \\
        & Avg & 4.79 & 6.60 & 7.59 & 8.00 & 6.7 \\
        \midrule
        
        \rowcolor{cyan!4}
        & Min & 2   & 2   & 2   & 2   & 2.0 \\
        \rowcolor{cyan!4}
        PRBI (Ours) & Max & \textcolor{cyan!60!black}{\text{2}}  
              & \textcolor{cyan!60!black}{\text{4}}  
              & \textcolor{cyan!60!black}{\text{6}}  
              & \textcolor{cyan!60!black}{\text{8}}  
              & \textcolor{cyan!60!black}{\text{5.0}} \\
        \rowcolor{cyan!4}
        & Avg & \textcolor{cyan!60!black}{\text{2.00}} 
              & \textcolor{cyan!60!black}{\text{2.35}} 
              & \textcolor{cyan!60!black}{\text{2.61}} 
              & \textcolor{cyan!60!black}{\text{2.86}} 
              & \textcolor{cyan!60!black}{\text{2.5}} \\
        \bottomrule
    \end{tabular}
    \vspace{-5mm}
\end{table}

\textbf{Comparison with Counterparts.}~\cref{tab:vercount} reports the verification cost per frame for $n=5$ under varying attacker ratios. PRBI achieves the lowest average cost (2.5 verifications per frame) and remains stable between 2 and 3 across all attack intensities, demonstrating strong robustness to adversarial variation. When both groups contain attackers in the first abnormal frame, PRBI employs a divide-and-conquer query strategy. Although this may slightly increase the peak cost, the search terminates once a benign subset is found, incurring minimal delay. Overall, the maximum verification count of PRBI averages 5.0, significantly lower than ROBOSAC (30.3) and PASAC (7.5).

\begin{table*}[t]\fontsize{8pt}{10pt}\selectfont
    \centering
    \caption{Detection performance when $n=5$ and $k=2$ under different fusion strategies. \textit{Lower-Bound} denotes single-vehicle perception, while \textit{Upper-Bound} represents collaborative perception without attacks. `$w$' indicates `with'. Best results are marked in \textcolor{cyan!60!black}{\text{blue}}, and comparisons with baseline methods demonstrate the average defense performance across test datasets.}
    \vspace{-2mm}
    \label{tab:performance}
    \setlength{\tabcolsep}{1pt}
    \begin{tabular}{@{}l|cc|cc|cc|cc|cc@{}}
        \toprule
        \multirow{2}{*}{\textbf{Methods}} 
        & \multicolumn{2}{c|}{\textbf{Mean Fusion}} 
        & \multicolumn{2}{c|}{\textbf{Max Fusion}} 
        & \multicolumn{2}{c|}{\textbf{Sum Fusion}} 
        & \multicolumn{2}{c|}{\textbf{V2VNet~\cite{wang2020v2vnet}}} 
        & \multicolumn{2}{c}{\textbf{DiscoNet~\cite{li2021learning}}} \\ 
        \cmidrule(lr){2-11}
        & \textbf{AP@0.5$\uparrow$} & \textbf{AP@0.7$\uparrow$}
        & \textbf{AP@0.5$\uparrow$} & \textbf{AP@0.7$\uparrow$}
        & \textbf{AP@0.5$\uparrow$} & \textbf{AP@0.7$\uparrow$}
        & \textbf{AP@0.5$\uparrow$} & \textbf{AP@0.7$\uparrow$}
        & \textbf{AP@0.5$\uparrow$} & \textbf{AP@0.7$\uparrow$} \\
        \midrule
        Upper-Bound & 79.57 & 75.34 & 78.62 & 75.07 & 77.88 & 74.38 & 80.73 & 78.35  & 80.52 & 76.72\\
        \midrule
        Attack $w$ PGD  & 16.51 & 14.60 & 27.16 & 26.45 & 13.65 & 7.31 & 17.02 & 14.53 & 20.97 & 19.54 \\
        \rowcolor{cyan!4}
        PRBI against PGD  & $\text{65.36}_{\textcolor{blue}{+48.85}}$ & $\text{61.44}_{\textcolor{blue}{+46.84}}$ & $\text{64.89}_{\textcolor{blue}{+37.73}}$ & $\text{62.02}_{\textcolor{blue}{+35.57}}$ & $\text{70.78}_{\textcolor{blue}{+57.13}}$ & $\text{67.67}_{\textcolor{blue}{+60.36}}$ & $\text{68.93}_{\textcolor{blue}{+51.91}}$ & $\text{63.82}_{\textcolor{blue}{+49.29}}$ & $\text{69.18}_{\textcolor{blue}{+48.21}}$ & $\text{65.68}_{\textcolor{blue}{+46.14}}$ \\
        \midrule
        Attack $w$ BIM  & 13.67 & 9.08 & 33.54 & 32.71 & 4.56 & 4.13 & 13.51 & 11.69 & 21.02 & 18.53 \\
        \rowcolor{cyan!4}
        PRBI against BIM  & $\text{66.37}_{\textcolor{blue}{+52.70}}$ & $\text{60.72}_{\textcolor{blue}{+51.64}}$ & $\text{65.87}_{\textcolor{blue}{+32.33}}$ & $\text{63.30}_{\textcolor{blue}{+30.59}}$ & $\text{70.81}_{\textcolor{blue}{+66.25}}$ & $\text{67.65}_{\textcolor{blue}{+63.52}}$ & $\text{68.76}_{\textcolor{blue}{+55.25}}$ & $\text{64.88}_{\textcolor{blue}{+53.19}}$ & $\text{66.17}_{\textcolor{blue}{+45.15}}$ & $\text{63.27}_{\textcolor{blue}{+44.74}}$ \\
        \midrule
        Attack $w$ C\&W  & 7.51& 3.42 & 25.99 & 24.88 & 8.52 & 7.12 & 10.68 & 6.04 & 14.90 & 10.93 \\
        \rowcolor{cyan!4}
        PRBI against C\&W  & $\text{70.58}_{\textcolor{blue}{+63.07}}$ & $\text{68.49}_{\textcolor{blue}{+65.07}}$ & $\text{66.59}_{\textcolor{blue}{+40.60}}$ & $\text{61.91}_{\textcolor{blue}{+37.03}}$ & $\text{74.13}_{\textcolor{blue}{+65.61}}$ & $\text{71.16}_{\textcolor{blue}{+64.04}}$ & $\text{71.87}_{\textcolor{blue}{+61.19}}$ & $\text{68.54}_{\textcolor{blue}{+62.50}}$ & $\text{72.98}_{\textcolor{blue}{+58.08}}$ & $\text{69.06}_{\textcolor{blue}{+58.13}}$ \\
        \midrule
        Lower-Bound  & 56.35& 52.89 & 56.35 & 52.89 & 56.35 & 52.89 & 56.35 & 52.89 & 56.35 & 52.89 \\
        ROBOSAC~\cite{li2023among}  & ${\text{66.17}}_{\textcolor{blue}{+9.82}}$& ${\text{63.05}}_{\textcolor{blue}{+10.16}}$ & ${\text{63.58}}_{\textcolor{blue}{+7.23}}$ & ${\text{57.36}}_{\textcolor{blue}{+4.46}}$ & ${\text{64.07}}_{\textcolor{blue}{+7.72}}$ & ${\text{59.61}}_{\textcolor{blue}{+6.72}}$ & ${\text{64.13}}_{\textcolor{blue}{+7.78}}$ & ${\text{61.01}}_{\textcolor{blue}{+8.12}}$ & ${\text{65.29}}_{\textcolor{blue}{+8.94}}$ & ${\text{63.87}}_{\textcolor{blue}{+10.98}}$ \\
        PASAC~\cite{hu2025cp}  & ${\text{64.38}}_{\textcolor{blue}{+8.03}}$& ${\text{60.07}}_{\textcolor{blue}{+7.18}}$ & $\textcolor{cyan!60!black}{\text{65.96}}_{\textcolor{blue}{+9.61}}$ & ${\text{62.29}}_{\textcolor{blue}{+9.40}}$ & ${\text{67.98}}_{\textcolor{blue}{+11.63}}$ & ${\text{63.60}}_{\textcolor{blue}{+10.71}}$ & ${\text{68.39}}_{\textcolor{blue}{+12.04}}$ & ${\text{64.73}}_{\textcolor{blue}{+11.83}}$ & ${\text{69.08}}_{\textcolor{blue}{+12.73}}$ & ${\text{64.54}}_{\textcolor{blue}{+11.65}}$\\
        \rowcolor{cyan!4}
        PRBI (Ours) & $\textcolor{cyan!60!black}{\text{67.44}}_{\textcolor{blue}{+11.09}}$& $\textcolor{cyan!60!black}{\text{63.55}}_{\textcolor{blue}{+10.66}}$ & ${\text{65.78}}_{\textcolor{blue}{+9.43}}$ & $\textcolor{cyan!60!black}{\text{62.41}}_{\textcolor{blue}{+9.52}}$ & $\textcolor{cyan!60!black}{\text{71.91}}_{\textcolor{blue}{+15.56}}$ & $\textcolor{cyan!60!black}{\text{68.83}}_{\textcolor{blue}{+15.94}}$ & $\textcolor{cyan!60!black}{\text{69.85}}_{\textcolor{blue}{+13.50}}$ & $\textcolor{cyan!60!black}{\text{65.75}}_{\textcolor{blue}{+12.86}}$ & $\textcolor{cyan!60!black}{\text{69.44}}_{\textcolor{blue}{+13.09}}$ & $\textcolor{cyan!60!black}{\text{66.00}}_{\textcolor{blue}{+13.11}}$ \\

        \bottomrule
    \end{tabular}
    \vspace{-2mm}
\end{table*}

\begin{figure}[!ht]
\centering
\includegraphics[width=0.35\textwidth]{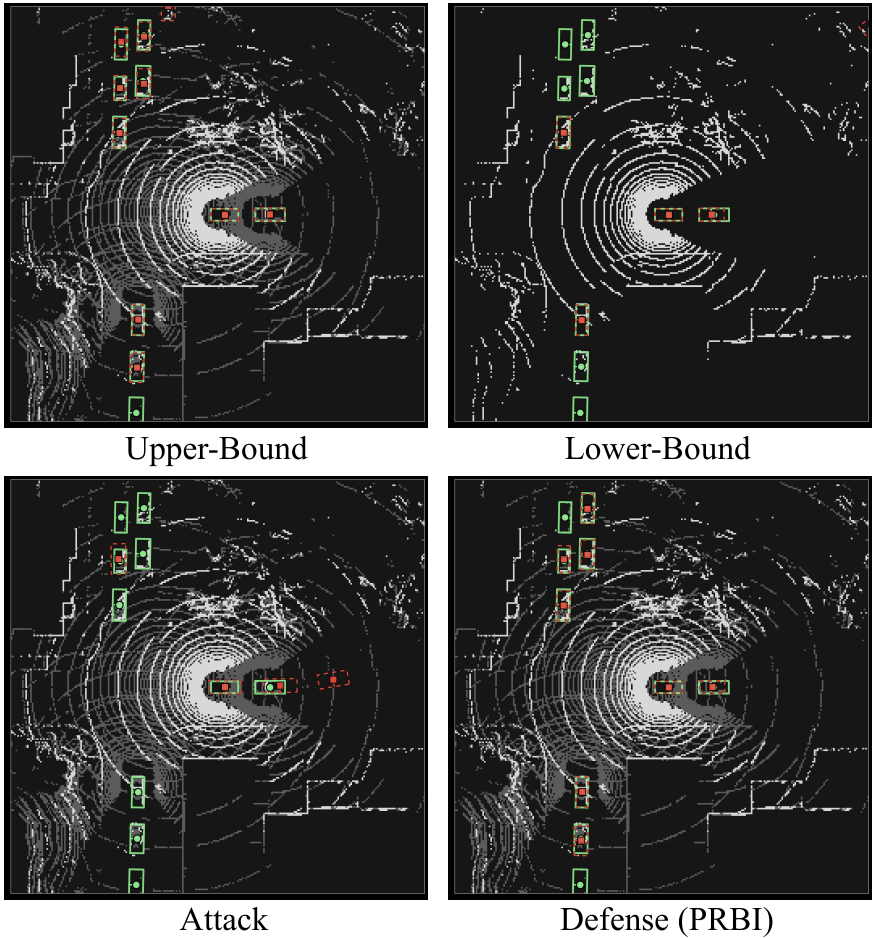}
\vspace{-3mm}
\caption{Visualization of defense performance with green boxes for ground truth and red boxes for predictions. }
\label{fig:vis}
\vspace{-5mm}
\end{figure}

\textbf{Detection Defense Performance.} 
\cref{tab:performance} summarizes results under different settings for $n=5$ and $k=2$. Under the V2VNet backbone, PRBI effectively restores performance, recovering 79.4\%, 81.0\%, and 86.9\% of AP loss against PGD, BIM, and C\&W, respectively. The defended results consistently exceed the Lower-Bound and approach the benign Upper-Bound. Moreover, PRBI maintains stable robustness across diverse fusion paradigms, achieving comparable recovery levels under all settings.
Compared with SOTA, PRBI achieves \text{1.3–3.5\%} higher AP than ROBOSAC and \text{1.0–4.0\%} higher than PASAC. Visualization in~\cref{fig:vis} further illustrate that PRBI successfully filters out adversarial vehicles and preserves accurate perception.

\begin{table}[htb!]\fontsize{8.5pt}{10pt}\selectfont
    \centering
    \caption{Impact of different test parameters on defense effect. \textit{Min/Max}: verification count range. \textit{ Avg. Min/Max}: average verification counts. \textit{Avg.}: overall average count. \textit{Avg. Frames}: average frames to convergence. \textit{ID Rate}: malicious vehicle identification rate. \textit{MC Rate}: benign vehicle misclassification rate.}
    \label{tab:testpara}
    \vspace{-2mm}
    \setlength{\tabcolsep}{3pt}
    \begin{tabular}{@{}l|cc|cc|c|c|cc@{}}
        \toprule
        \multirow{2}{*}{\textbf{Settings}} & \multicolumn{2}{c|}{\textbf{Count}} & \multicolumn{2}{c|}{\textbf{Avg. Count}} & \multirow{2}{*}{\textbf{Avg.}} & \textbf{Avg.} & \textbf{ID} & \textbf{MC} \\
        \cmidrule(lr){2-3} \cmidrule(lr){4-5}
        & \textbf{Min} & \textbf{Max} & \textbf{Min} & \textbf{Max} & & \textbf{Frames} & \textbf{Rate} & \textbf{Rate} \\
        \midrule
        \multicolumn{9}{c}{\textbf{\emph{Attacker Ratio}}} \\
        \midrule
        Ratio 80\% & 2 & 8 & 2.00 & 5.70 & 2.86 & 4.27 & 100\% & 0\% \\
        Ratio 60\% & 2 & 6 & 2.00 & 4.06 & 2.61 & 3.36 & 100\% & 0\% \\
        Ratio 40\% & 2 & 4 & 2.00 & 2.88 & 2.35 & 2.77 & 100\% & 6\% \\
        Ratio 20\% & 2 & 2 & 2.00 & 2.00 & 2.00 & 2.25 & 100\% & 0\% \\
        \midrule
        \multicolumn{9}{c}{\textbf{\emph{Confidence Level (60\% Attackers)}}} \\
        \midrule 
        Conf. 0.20 & 2 & 6 & 2.00 & 4.22 & 2.61 & 3.99 & 100\% & 0\% \\
        Conf. 0.15 & 2 & 6 & 2.00 & 4.12 & 2.62 & 3.51 & 100\% & 0\% \\
        Conf. 0.10 & 2 & 6 & 2.00 & 3.94 & 2.53 & 3.44 & 100\% & 0\% \\
        Conf. 0.05 & 2 & 6 & 2.00 & 4.26 & 2.67 & 3.40 & 100\% & 0\% \\
        Conf. 0.01 & 2 & 6 & 2.00 & 4.06 & 2.61 & 3.36 & 100\% & 0\% \\
        \midrule
        \multicolumn{9}{c}{\textbf{\emph{Window Size (80\% Attackers)}}} \\
        \midrule 
        Size 10 & 2 & 8 & 2.00 & 5.70 & 2.86 & 4.27 & 100\% & 0\% \\
        Size 8 & 2 & 8 & 2.00 & 6.16 & 2.91 & 4.60 & 100\% & 0\% \\
        Size 6 & 2 & 8 & 2.00 & 6.10 & 2.89 & 4.57 & 100\% & 0\% \\
        Size 4 & 2 & 8 & 2.00 & 6.34 & 2.88 & 4.89 & 100\% & 0\% \\
        \bottomrule
    \end{tabular}
    \vspace{-3mm}
\end{table}

\textbf{Varying Test Parameters.}
We evaluate the impact of the confidence level $\alpha$ and window size $w$ in the T-test on PRBI’s detection efficiency and accuracy, as summarized in~\cref{tab:testpara}. Specifically, we record the \textit{\textbf{Average Convergence Frames}}, the \textit{\textbf{Identification Rate}} of malicious vehicles, and the \textit{\textbf{Misclassification Rate}} of benign ones. Results show that PRBI converges within approximately \textit{four} frames on average while maintaining a 100\% identification rate for malicious vehicles. When the attacker ratio reaches 0.4, a slight performance drop (6\% misclassification) occurs because the test statistic $m$ stabilizes prematurely before all benign samples are included, causing the T-test to signal early convergence and misclassify one benign vehicle as malicious. Nevertheless, the malicious identification rate consistently remains 100\%. Moreover, sensitivity analysis on $\alpha$ and $w$ further supports this observation: reducing $\alpha$ accelerates convergence (frames decrease from 3.99 to 3.36) without degrading accuracy, while larger $w$ yields smoother convergence. Overall, PRBI demonstrates a robust balance between efficiency and reliability across all tested configurations. Detailed results are presented in the appendix.

\begin{figure}[htb!]
  \centering
  
  \begin{subfigure}[b]{0.48\columnwidth}
    \centering
    \includegraphics[width=\textwidth]{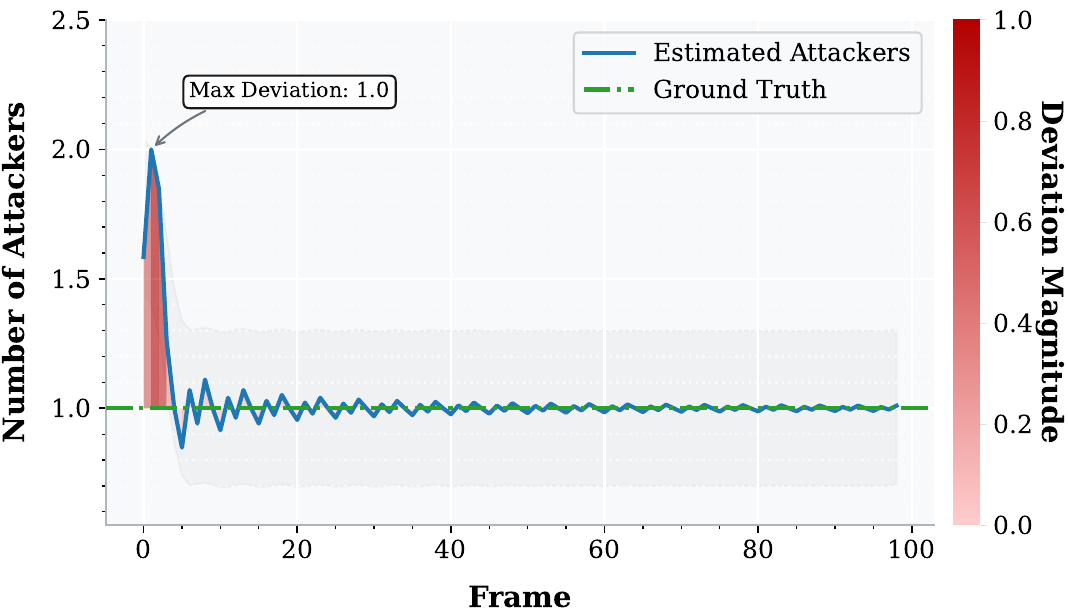}
    \caption{$k=1$ \& Floor}
  \end{subfigure}%
  \hfill
  \begin{subfigure}[b]{0.48\columnwidth}
    \centering
    \includegraphics[width=\textwidth]{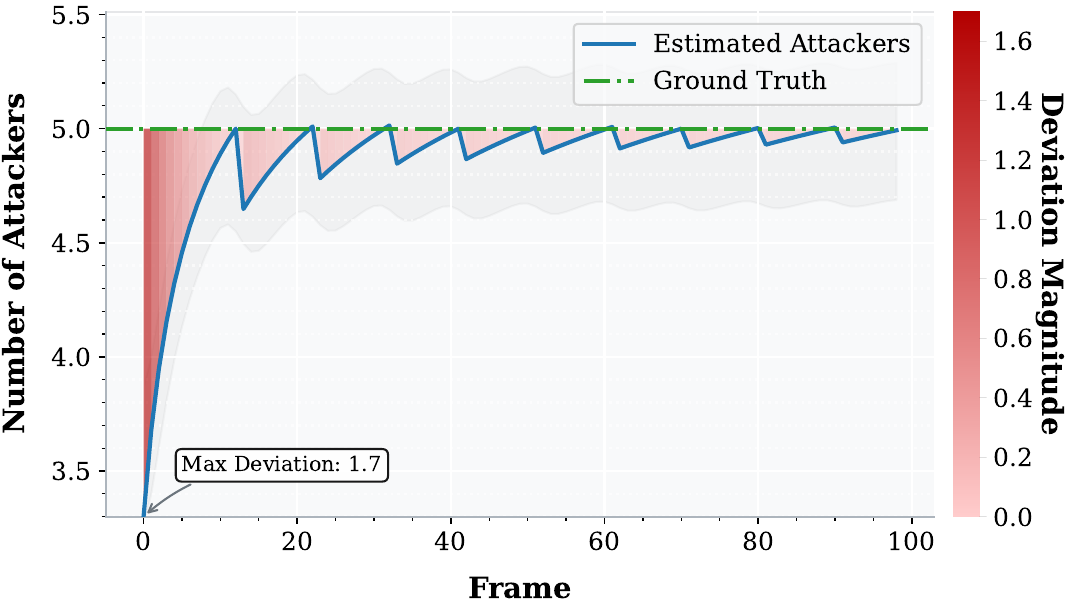}
    \caption{$k=5$ \& Floor}
  \end{subfigure}
  \begin{subfigure}[b]{0.48\columnwidth}
    \centering
    \includegraphics[width=\textwidth]{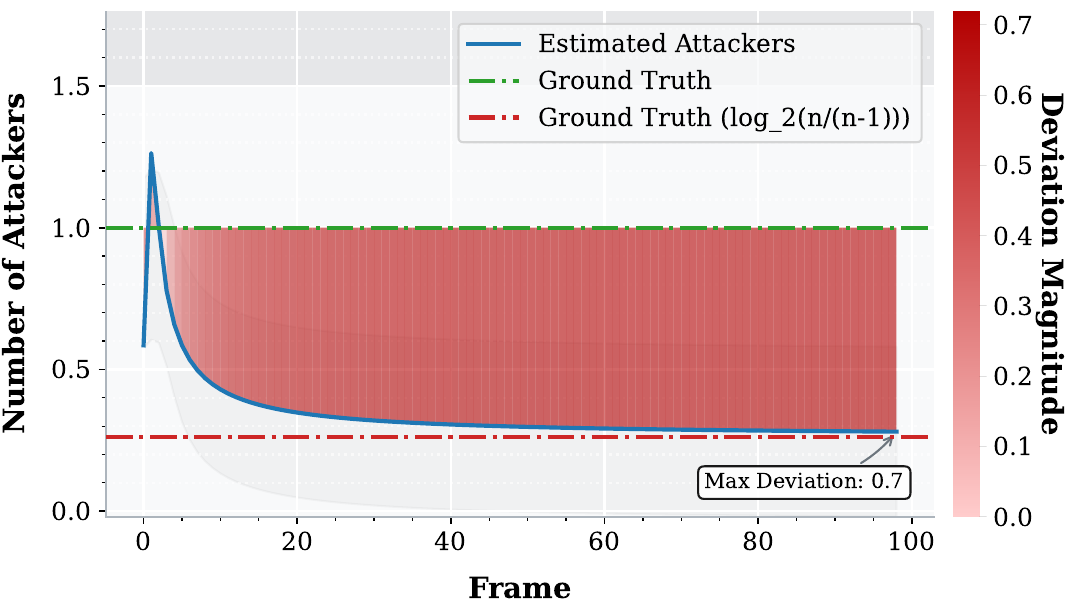}
    \caption{$k=1$ \& Ceil}
  \end{subfigure}%
  \hfill
  \begin{subfigure}[b]{0.48\columnwidth}
    \centering
    \includegraphics[width=\textwidth]{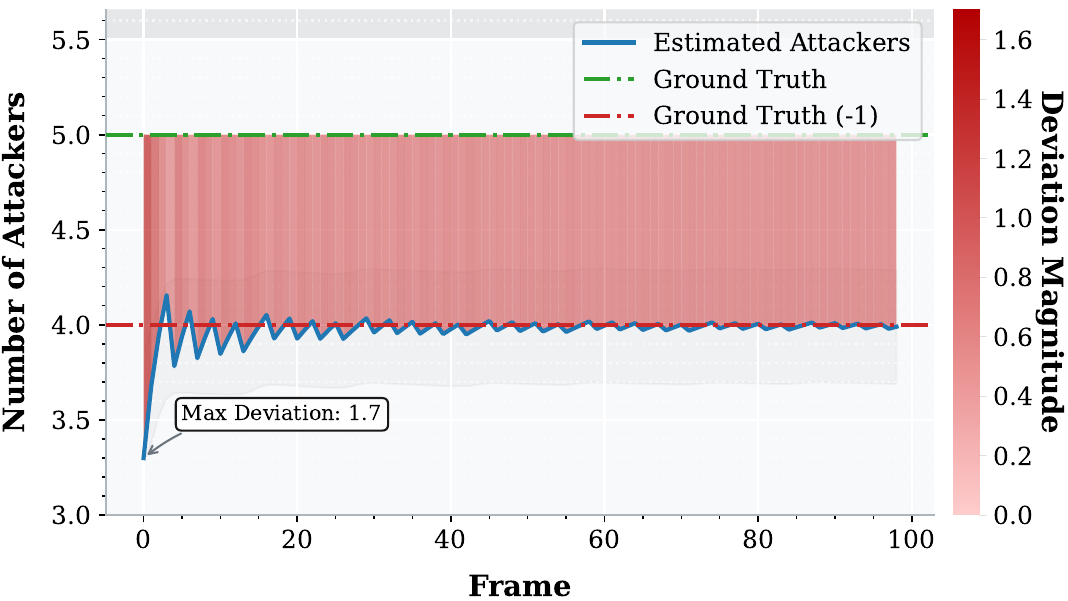}
    \caption{$k=5$ \& Ceil}
  \end{subfigure}
  \begin{subfigure}[b]{0.48\columnwidth}
    \centering
    \includegraphics[width=\textwidth]{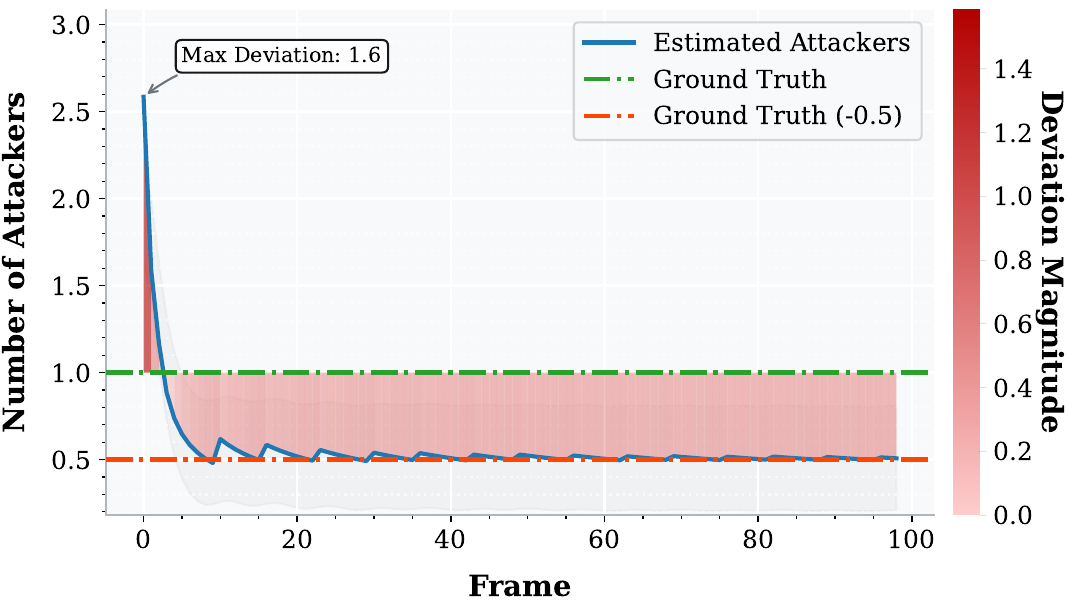}
    \caption{$k=1$ \& Round}
  \end{subfigure}%
  \hfill
  \begin{subfigure}[b]{0.48\columnwidth}
    \centering
    \includegraphics[width=\textwidth]{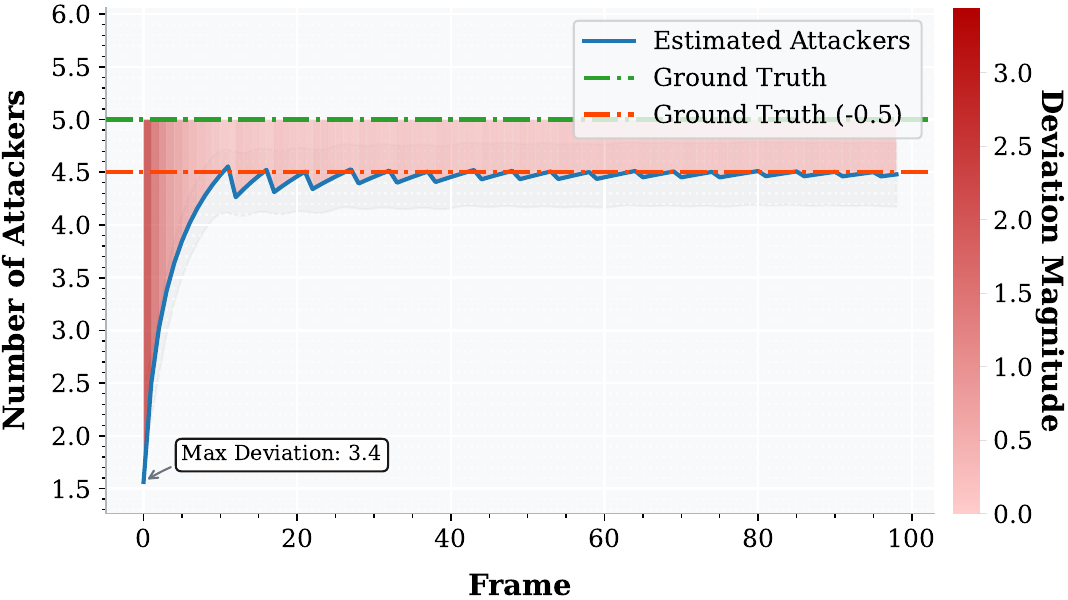}
    \caption{$k=5$ \& Round}
  \end{subfigure}

  \vspace{-1mm}
  \caption{Convergence behavior of $m$ with total 6 vehicles under different rounding strategies. The blue solid line denotes the value of $m$ calculated using~\cref{eq:m_compute}, the green dashed line marks the true number of attackers $k$, and the red dashed line represents the theoretically derived convergence value in~\cref{thm:rounding}.}
  \vspace{-2mm}
  \label{fig:conplot}
\end{figure}

\begin{figure}[htb!]
  \centering
  
  \begin{subfigure}[b]{0.48\columnwidth}
    \includegraphics[width=\textwidth]{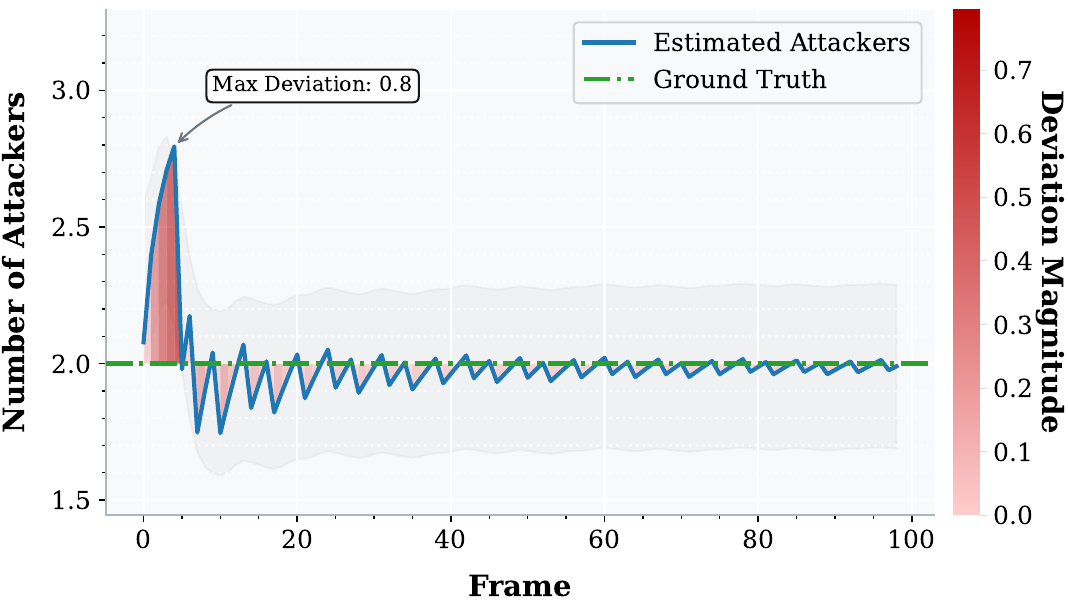}
    \caption{$n=20, k=2$}
    \label{fig:sub_202}
  \end{subfigure}
  \hfill
  \begin{subfigure}[b]{0.48\columnwidth}
    \includegraphics[width=\textwidth]{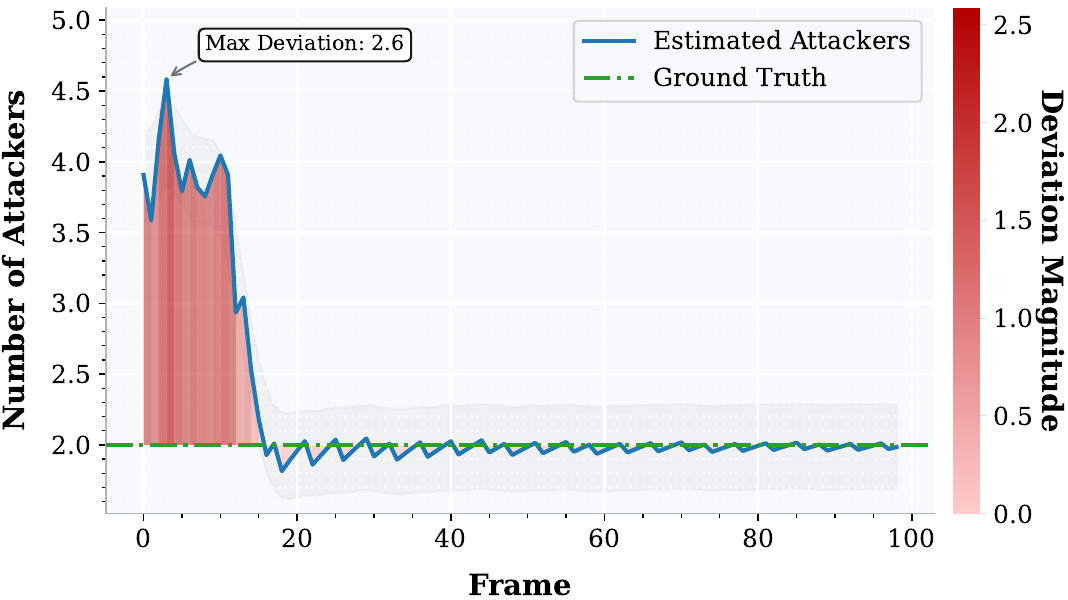}
    \caption{$n=30, k=2$}
    \label{fig:sub_302}
  \end{subfigure}
  \vspace{-2mm}
  \begin{subfigure}[b]{0.48\columnwidth}
    \includegraphics[width=\textwidth]{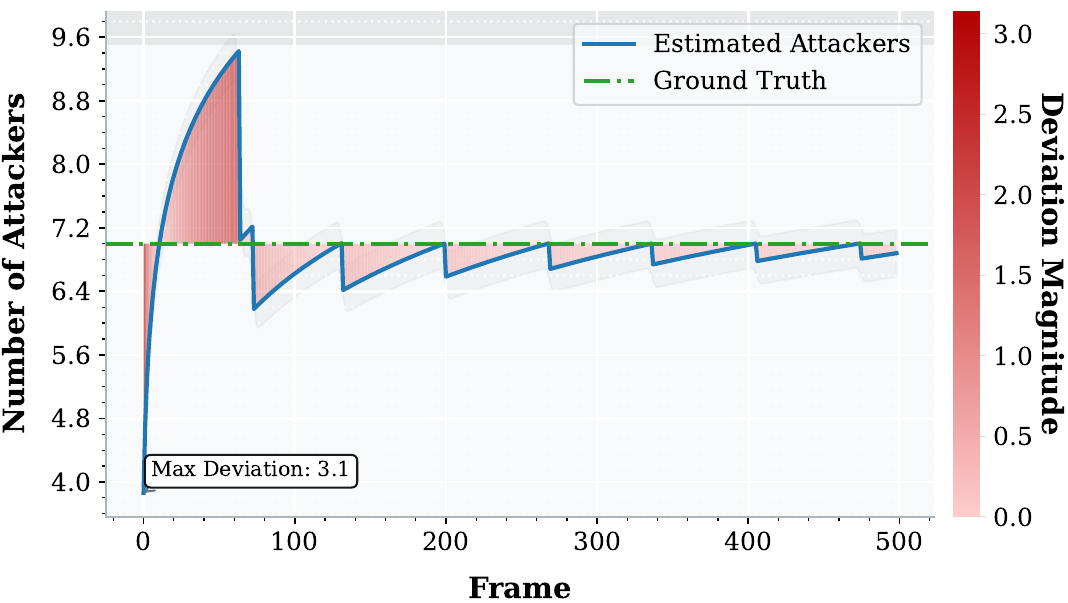}
    \caption{$n=15$, attacker ratio is 50\%}
    \label{fig:sub_1550}
  \end{subfigure}
  \hfill
  \begin{subfigure}[b]{0.48\columnwidth}
    \includegraphics[width=\textwidth]{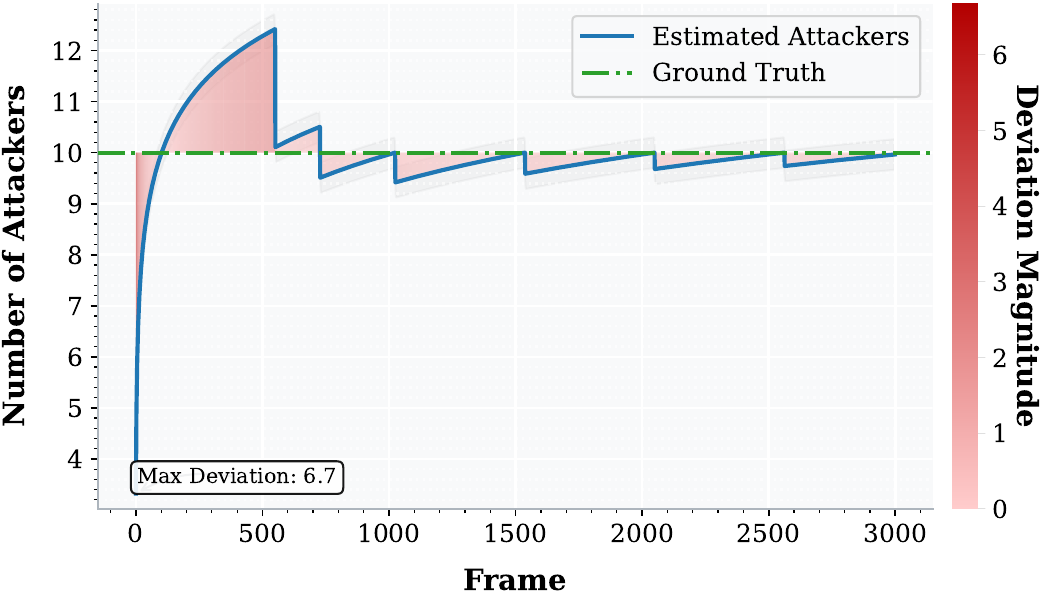}
    \caption{$n=20$, attacker ratio is 50\%}
    \label{fig:sub_2050}
  \end{subfigure}
  \caption{The convergence characteristics of PRBI's frame-by-frame estimation of the calculated malicious vehicles number $m$.}
  \vspace{-2mm}
  \label{fig:conplot_ratio}
  \label{fig:conplot_chebubian}
\end{figure}

\subsection{Algorithm Convergence}
To intuitively verify the convergence of PRBI (\cref{sec:proof}), we record the evolution of $m$ under three rounding strategies with six vehicles and plot the results in~\cref{fig:conplot}. When using floor rounding, $m$ consistently converges to the true attacker count $k$, in line with~\cref{thm:convergence}. For rounding-to-nearest and ceiling modes, $m$ converges to $k - 0.5$ and $k - 1$, respectively. Notably, when $k=1$, the ceiling mode yields $m \approx \log_2(n/(n-1)) \approx 0.263$, perfectly matching the theoretical result in~\cref{thm:rounding}. Across all settings, the deviation between estimated $m$ and $k$ never exceeds 2, confirming the numerical stability induced by the logarithmic form. 

To further analyze convergence behavior, we examine two scenarios with varying vehicle counts. In~\cref{fig:sub_202,fig:sub_302}, with a fixed number of attackers, larger $n$ slows convergence since a greater denominator increment is needed in~\cref{eq:m_compute} to satisfy the logarithmic condition. When the attacker ratio is fixed, both $n$ and $k$ increase, causing the logarithmic term to grow rapidly and convergence to slow down. As illustrated in~\cref{fig:sub_1550,fig:sub_2050}, convergence takes about 100 frames for $n=15$ and up to 500 frames for $n=20$.

\subsection{Probability Trend of Vehicles}
\begin{figure}[t]
  \centering
  \begin{subfigure}[b]{0.23\textwidth}
    \includegraphics[width=\textwidth]{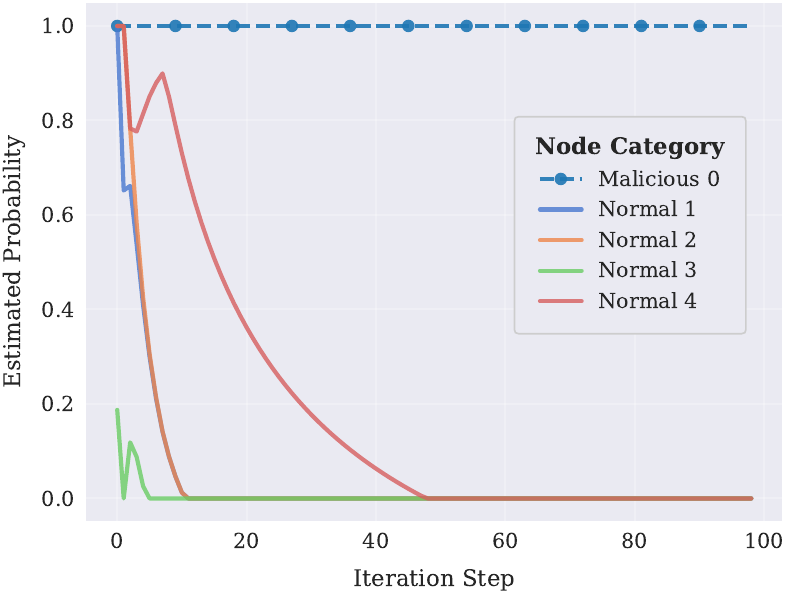}
    \caption{$k=1$}
  \end{subfigure}\hspace{0.01\textwidth}
  \begin{subfigure}[b]{0.23\textwidth}
    \includegraphics[width=\textwidth]{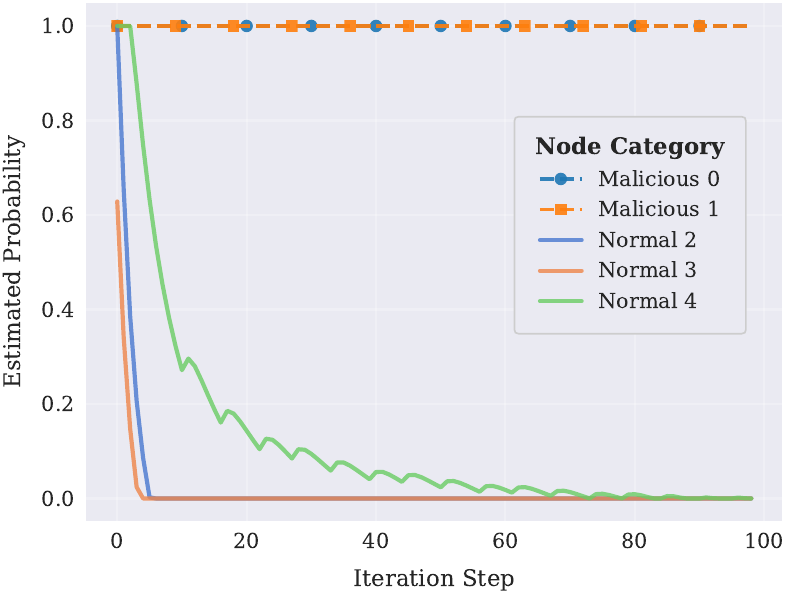}
    \caption{$k=2$}
  \end{subfigure}\hspace{0.01\textwidth}
  \begin{subfigure}[b]{0.23\textwidth}
    \includegraphics[width=\textwidth]{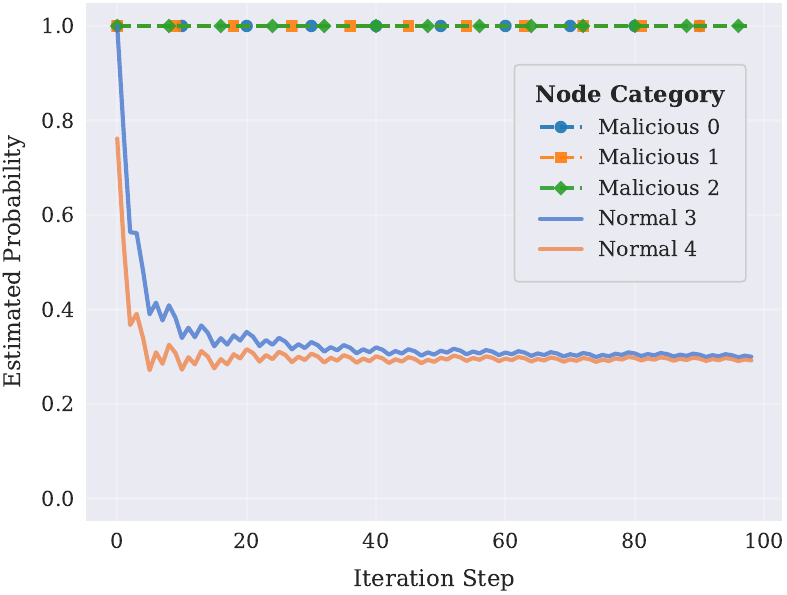}
    \caption{$k=3$}
  \end{subfigure}\hspace{0.01\textwidth}
  \begin{subfigure}[b]{0.23\textwidth}
    \includegraphics[width=\textwidth]{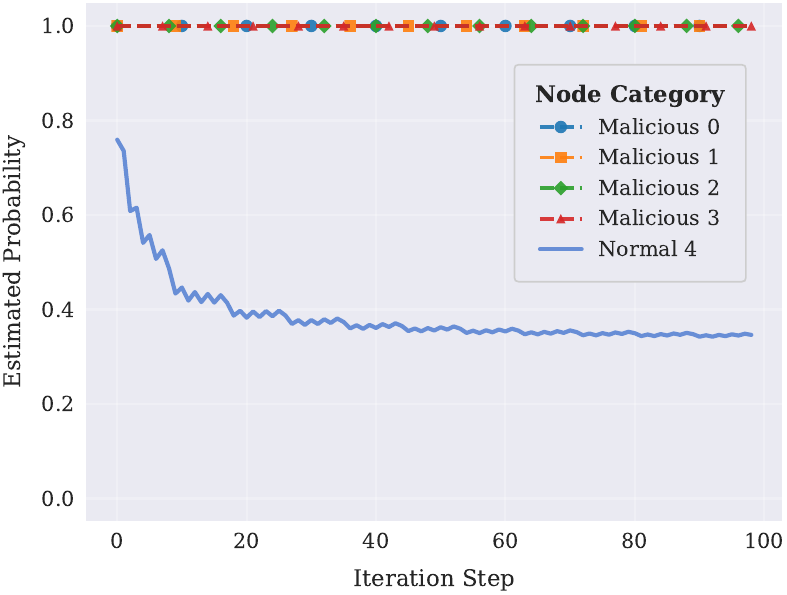}
    \caption{$k=4$}
  \end{subfigure}
  \vspace{-5mm}
  \caption{The probability of malicious behavior for all vehicles under varying numbers of attackers.}
  \vspace{-3mm}
  \label{fig:conplot_prob}
\end{figure}

To analyze how the benign probability of each vehicle evolves during defense, we set $n=5$ and compute the complement of $P_{\text{benign}}$ to obtain the malicious probability, as shown in~\cref{fig:conplot_prob}. Since groups containing attackers are more likely to trigger abnormal detections, the normal detection count $\beta_j$ of a malicious vehicle remains zero, causing its benign probability to vanish. Consequently, the malicious probabilities of all attackers rapidly reach 100\%.

%% file: sec/7_conclusion.tex
\section{Conclusion}
This paper tackles the challenge of defending against adversarial attacks in fully untrusted-vehicle CP by proposing the PRBI framework, which integrates adjacent-frame similarity analysis with pseudo-random Bayesian inference. By exploiting stability differences in perception results across adjacent frames, PRBI eliminates reliance on a trusted ego vehicle. It combines group-based validation and probabilistic inference to accurately identify malicious vehicles with only two additional verifications per frame. Theoretical analysis demonstrates the convergence of the estimated number of attackers, and comprehensive experiments further verify PRBI’s significant advantages in defense efficiency and performance. These results highlight PRBI as an effective and resource-efficient defense for CP security in autonomous driving. 

\noindent {\small \textbf{Acknowledgement.} This work was supported by the National Natural Science Foundation of China (62441237, U24A20336), the Fundamental Research Funds for the Central Universities (2042025kf0055) and Wuhan City Joint Innovation Laboratory for Next-Generation Wireless Communication Industry Featuring Satellite-Terrestrial Integration under Grant 4050902040448.}

%% file: sec/X_suppl.tex
\clearpage
\setcounter{page}{1}
\maketitlesupplementary

\section{Hypothesis Verification Experiments}
\label{appendix:hypothesisverification}

To validate our assumption regarding the temporal stability of LiDAR-based perception, we conduct a dedicated hypothesis verification experiment using the V2X-Sim dataset~\cite{li2022v2x} within the collaborative perception framework. Our goal is to examine whether perception outputs across consecutive frames maintain consistently high similarity under normal conditions, while experiencing a significant drop in similarity under adversarial perturbations.

\begin{figure*}[t]
    \centering
    \includegraphics[width=0.8\textwidth]{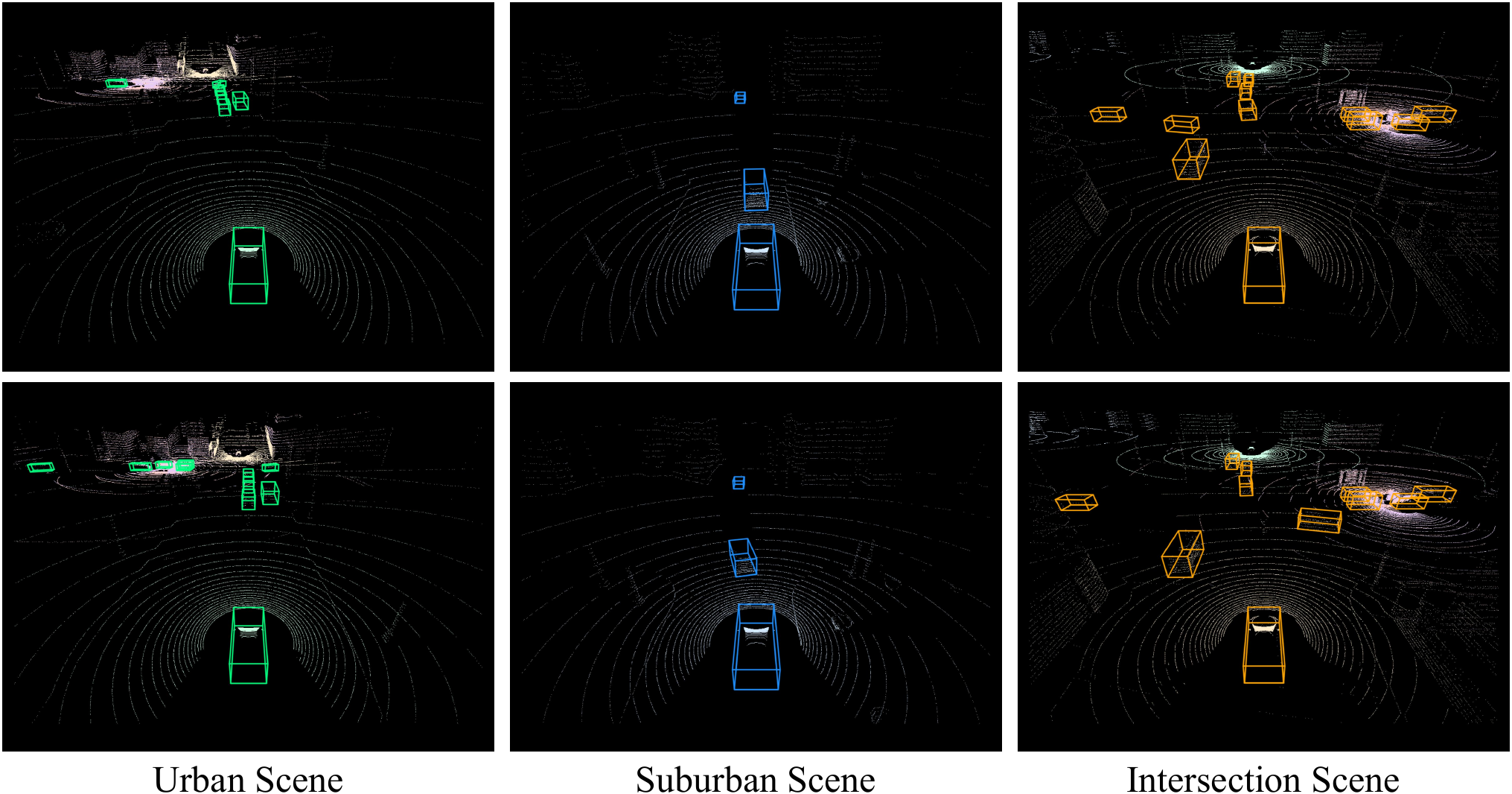}
    \caption{Example LiDAR point clouds illustrating three representative driving scenes in V2X-Sim dataset: urban, suburban, and intersection.}
    \label{fig:scene_radar}
\end{figure*}

\textbf{Experimental Setup.} We consider three representative types of driving scenes in V2X-Sim dataset: \textit{urban}, \textit{suburban}, and \textit{intersection}, as shown in~\cref{fig:scene_radar}, where we provide representative LiDAR point cloud visualizations for each scene type. For each category, we randomly select 5–10 scenes, each containing multiple consecutive frames of multi-vehicle LiDAR data. Within each scene, we compute the Jaccard similarity between consecutive-frame detection outputs, using V2VNet~\cite{wang2020v2vnet} as the perception backbone. The evaluation is carried out for both:
\begin{itemize}[labelsep=0.5em, leftmargin=*]
    \item \textbf{Normal settings:} 
    \begin{enumerate}
        \item \textit{Upper-Bound:} All vehicles are collaborative and benign.
        \item \textit{Lower-Bound:} Only the ego vehicle's own perception is used (\textit{single-vehicle baseline}).
    \end{enumerate}
    \item \textbf{Adversarial settings:} All scenes re-evaluated under three widely used white-box attacks—BIM~\cite{kurakin2018adversarial}, C\&W~\cite{carlini2017towards}, and PGD~\cite{madry2017towards}. Perturbations are injected into the feature maps of randomly selected malicious vehicles, following standard attack parameters.
\end{itemize}

\begin{figure}[!th]
  \centering
  \begin{subfigure}[b]{0.22\textwidth}
    \includegraphics[width=\textwidth]{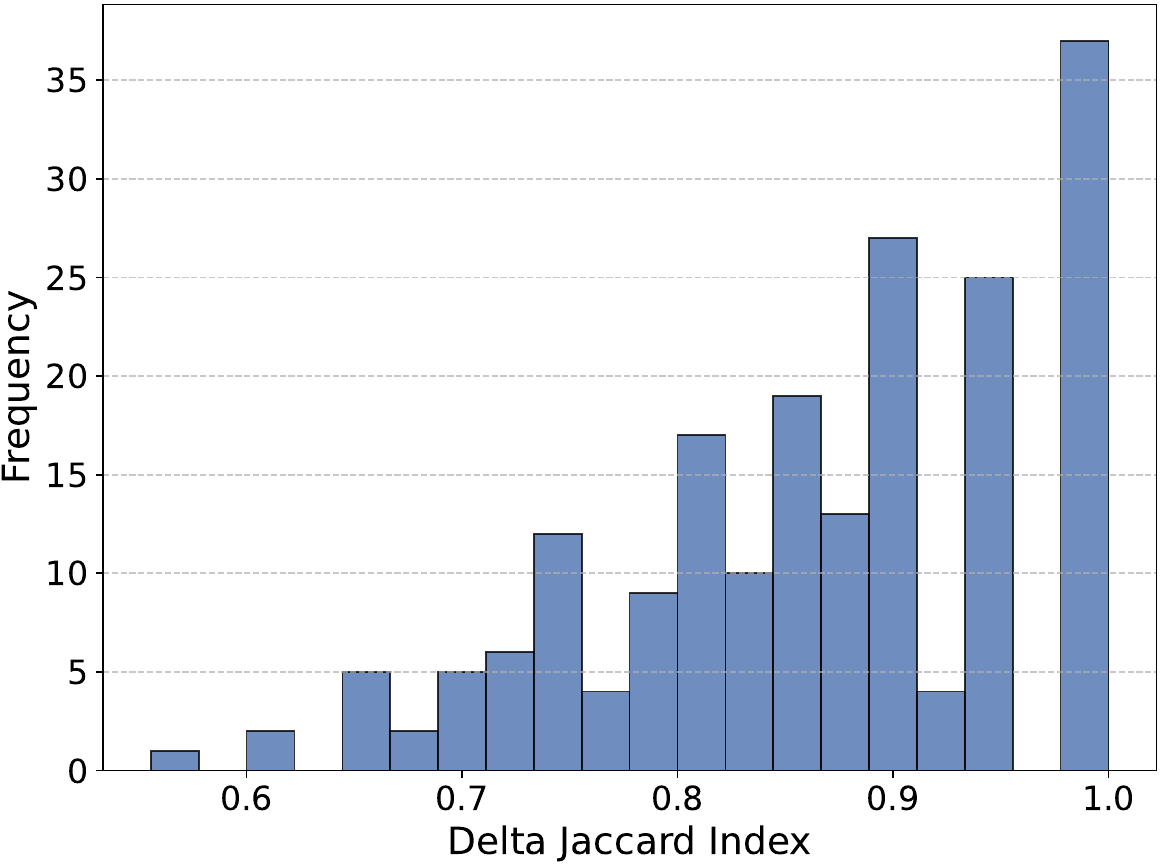}
    \caption{Upper-Bound}
  \end{subfigure}
  \begin{subfigure}[b]{0.22\textwidth}
    \includegraphics[width=\textwidth]{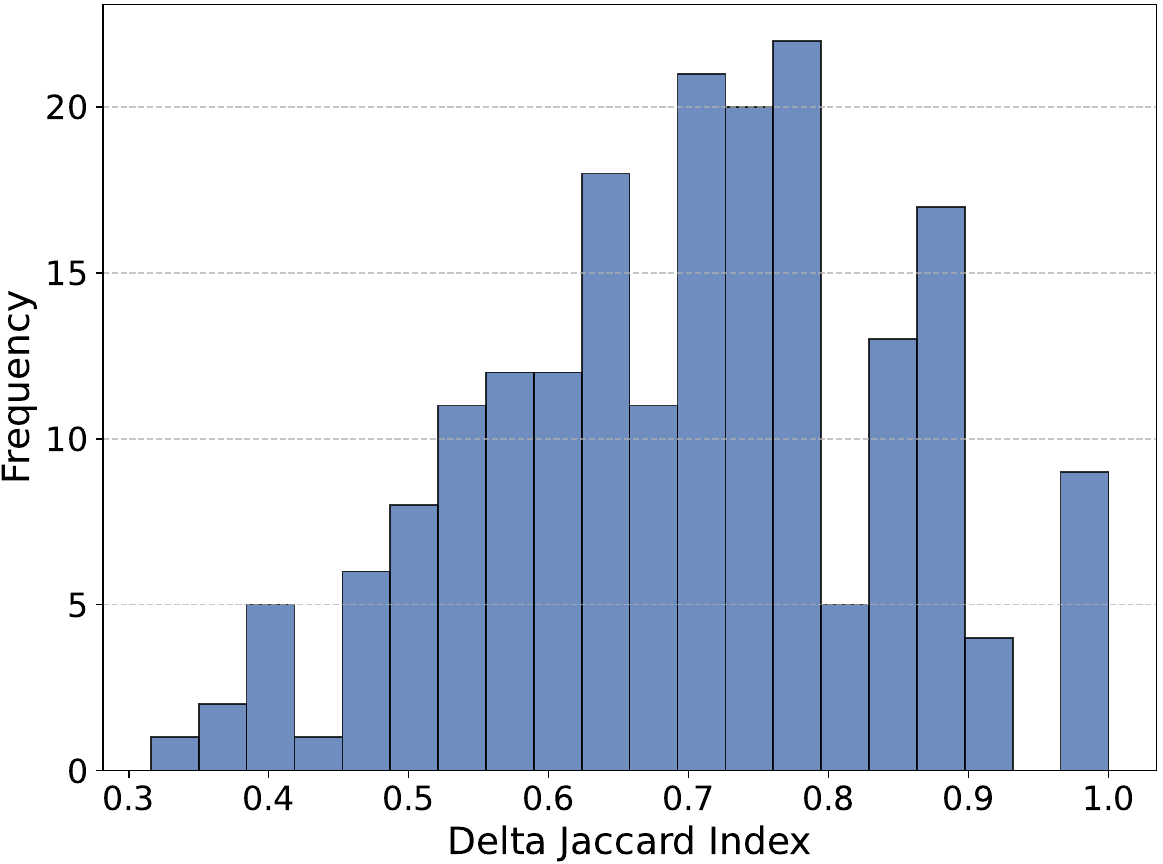}
    \caption{Lower-Bound}
  \end{subfigure}
  \begin{subfigure}[b]{0.22\textwidth}
    \includegraphics[width=\textwidth]{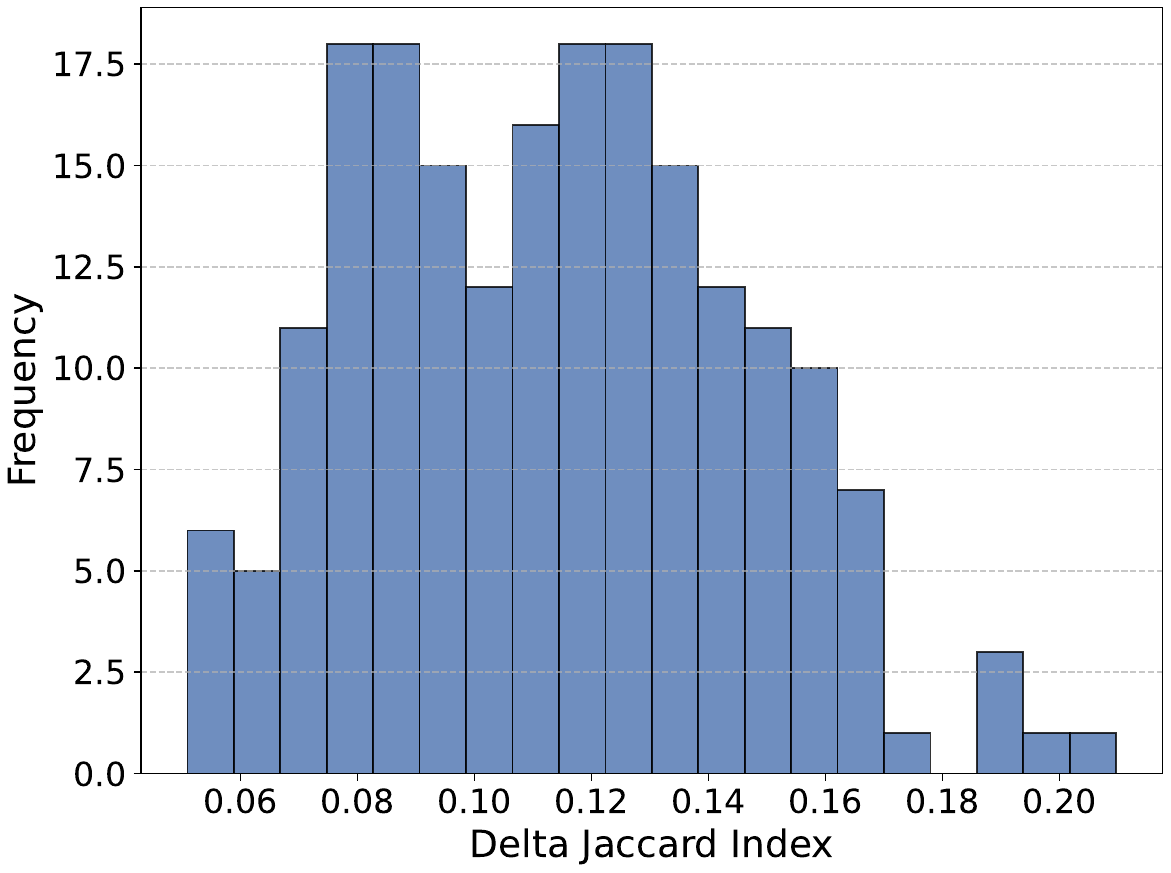}
    \caption{BIM Attack}
  \end{subfigure}
  \begin{subfigure}[b]{0.22\textwidth}
    \includegraphics[width=\textwidth]{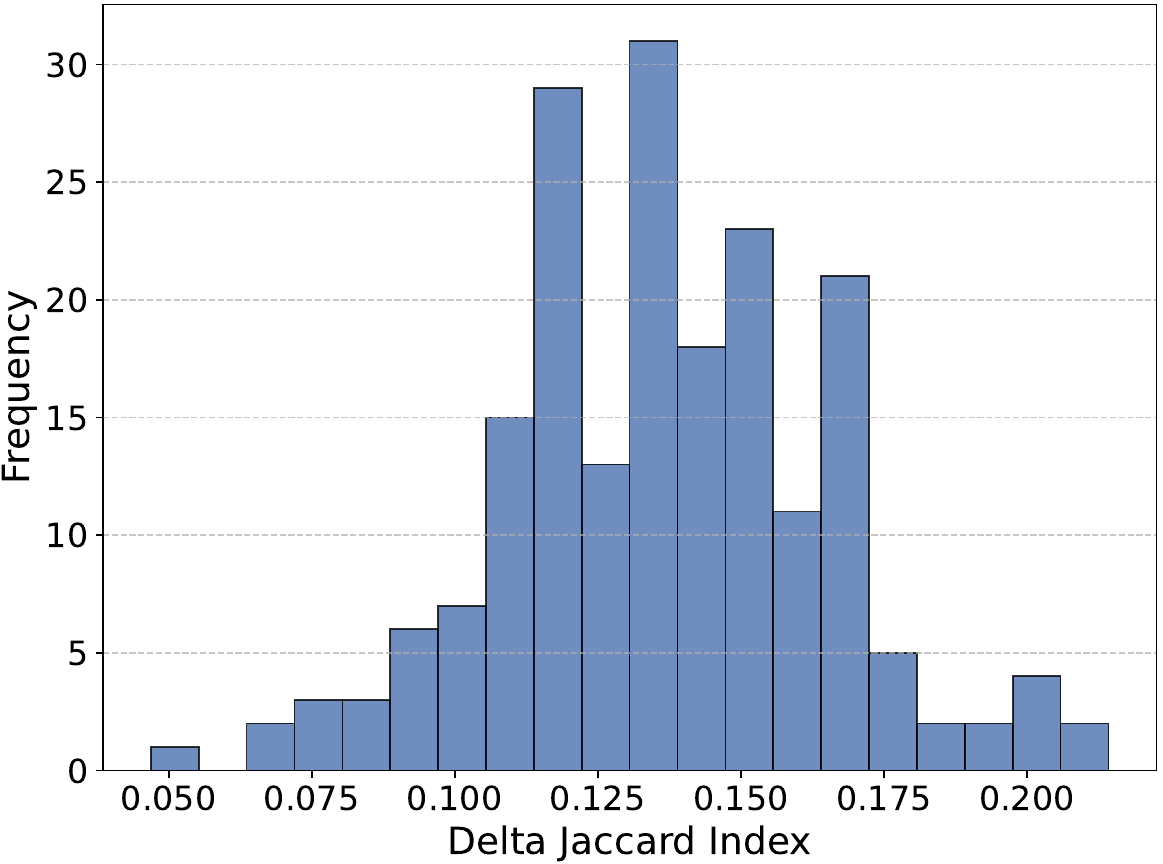}
    \caption{C\&W Attack}
  \end{subfigure}
  \begin{subfigure}[b]{0.22\textwidth}
    \includegraphics[width=\textwidth]{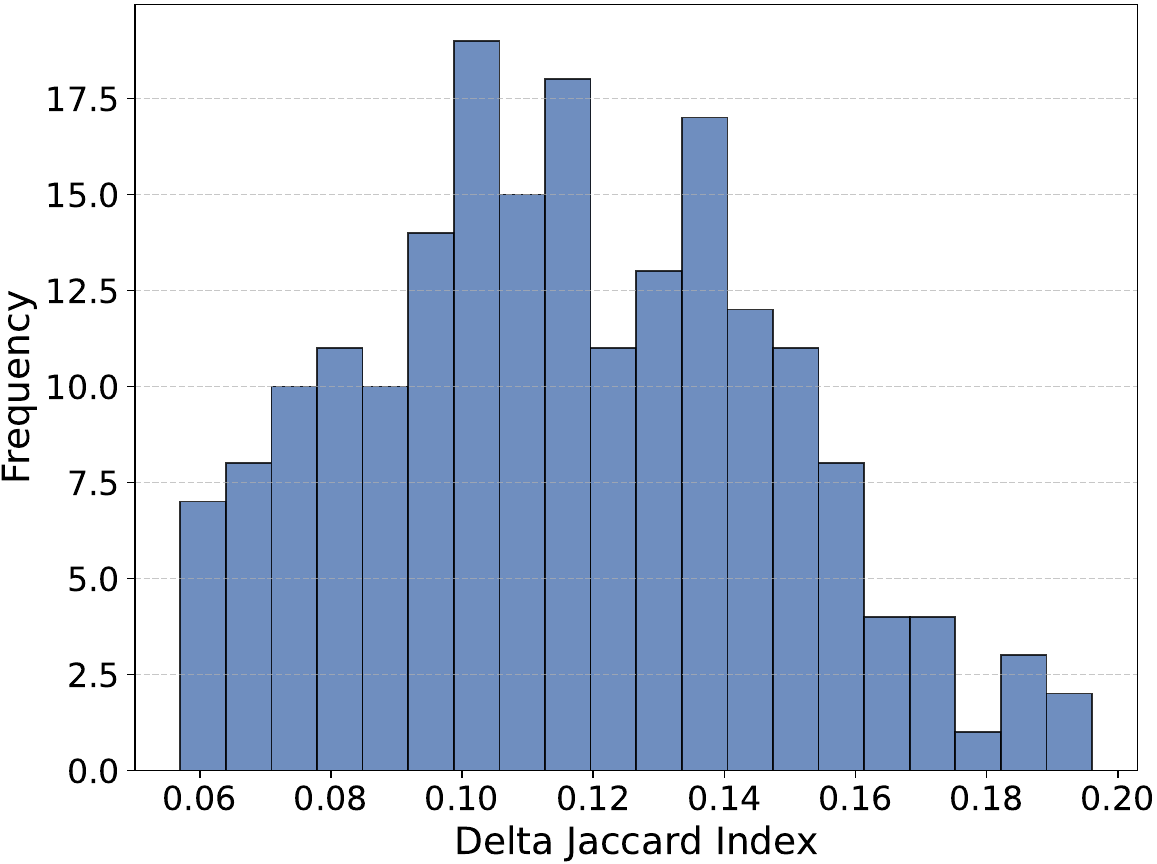}
    \caption{PGD Attack}
  \end{subfigure}
  \caption{
  Hypothesis verification experiments. We report the Jaccard similarity of detected object sets between consecutive frames for:
  \textit{(a)-(b)} normal conditions (Upper-Bound: all vehicles collaborate; Lower-Bound: ego-only perception), and
  \textit{(c)-(e)} adversarial conditions (BIM, C\&W, and PGD). 
  Plots are based on statistics from two randomly selected representative scenes for readability; trends are consistent across all tested scenes and scenarios (urban, suburban, intersection). 
  }
  \label{fig:jaccardindex}
\end{figure}

\textbf{Observations and Results.} For~\cref{fig:jaccardindex}, we visualize distribution statistics from \emph{two randomly picked scenes (scene 4 \& 78)} out of the entire test scenarios. This sub-selection is purely for clarity and readability of the plots; the trends are representative of the complete experimental results across all selected scenes.~\cref{fig:jaccardindex} clearly illustrates an evident separation between normal and adversarial regimes:
\begin{itemize}[labelsep=0.5em, leftmargin=*]
    \item Under \textbf{normal conditions} (both Upper- and Lower-Bound), inter-frame similarity consistently clusters around $\mathbf{0.8}$, indicating strong temporal stability and spatial continuity of perception outputs.
    \item Under \textbf{adversarial attacks} (BIM, C\&W, PGD), similarity values drop sharply, \emph{often well below 0.3, and peaking only around 0.2}, reflecting the severe disruption of temporal consistency caused by injected perturbations.
\end{itemize}

These results hold across scene types and independent random samplings, confirming that the temporal stability signal is strong and scenario-agnostic. 

\textbf{Conclusion of Hypothesis Verification.} The persistent and large margin between benign and adversarial similarity distributions validates our assumption: temporal similarity between successive LiDAR perception outputs is consistently high under benign conditions and substantially degraded under attack. This robust gap can serve as a reliable self-referential signal, forming a sound foundation for our proposed frame-wise self-supervision mechanism, without the need to assume a trusted ego vehicle.

\begin{algorithm}[!ht]
\caption{Pseudo-Random Bayesian Inference}\label{alg:prbi}
\renewcommand{\algorithmicrequire}{\textbf{Input:}}
\renewcommand{\algorithmicensure}{\textbf{Output:}}
\begin{algorithmic}[1]

\Require Number of CAVs $n$; Detection count arrays $\mathbf{c}_{\text{normal}}$, $\mathbf{c}_{\text{abnormal}}$; Similarity threshold $\epsilon$; Hypothesis testing window size $w$.
\Ensure Malicious vehicles: $Attackers = \{id_1, \dots, id_k\}$


\State $\mathbf{c}_{\text{normal}} \gets \mathbf{0}$, $\mathbf{c}_{\text{abnormal}} \gets \mathbf{0}$, $N \gets 0$ 
\State $m \gets n - 1$, $P_{\text{benign}} \gets \mathbf{0}$ 
\State $W \gets \text{Queue}()$, $convergence \gets \textbf{False}$

\While{system running, $i \gets i + 1$}

    \State $D_1^i \gets \operatorname{Perception}(F_1^i, \dots, F_n^i)$ 

    \If{$i = 0$}
        \State $D_{\text{ref}} \gets D_1^i$ \Comment{Initialize reference frame}
        \State $D_1^i$ as \textit{PerceptionOutput}
        \State \textbf{continue}
    \EndIf

    \If{$\operatorname{Jaccard}(D_1^i, D_{\text{ref}}) < \epsilon$ \textbf{and} $convergence = \textbf{False}$}

        \State \textcolor{brown}{$\triangleright$ \textit{Soft Sampling}}
        \State $Group_1 \gets \text{argsort}(P_{\text{benign}})_{0:\left\lfloor m \right\rfloor}$  
        \State $Group_2 \gets \text{argsort}(P_{\text{benign}})_{\left\lfloor m \right\rfloor:n}$

        \State \textcolor{brown}{$\triangleright$  \textit{Consistency Validation}}

        \State $D_{\text{Group}_{1,2}} \gets \operatorname{Perception}(Group_{1,2})$ 
        \State Compare $\operatorname{Jaccard}(D_{\text{Group}_{1,2}}, D_{\text{ref}})$ with $\epsilon$

        \State Update $\mathbf{c}_{\text{normal}}$, $N$, and $\mathbf{c}_{\text{abnormal}}$

        \State \textcolor{brown}{$\triangleright$  \textit{Attacker Evaluation}}

        \State $m \gets \min\left(\log_2\left(\frac{n \cdot N}{\sum(\mathbf{c}_{\text{normal}})}\right), n - 1\right)$ 

        \For{$id = 1$ \textbf{to} $n$} 
            \State $P_{\text{benign}}[id] \gets \operatorname{Bayesian}(\mathbf{c}_{\text{nor}}, \mathbf{c}_{\text{abnor}}, N)$ 
        \EndFor

        \State $Attackers \gets \text{argsort}(P_{\text{benign}})_{0:\left\lfloor m \right\rceil}$

        \State \textcolor{brown}{$\triangleright$ \textit{Hypothesis Test}}

        \State $W.\text{push}(m)$ \textbf{with} size limit $w$ 

        \If{$\operatorname{T\text{-}test}(W)$ indicates convergence}
            \State $convergence \gets \textbf{True}$
            \State \textbf{Output} $Attackers$ 
        \Else
            \State $Collaborators \gets \{i \mid P_{\text{benign}}[i] \neq 0\}$
            \If{$Collaborators=\emptyset$ \textbf{and} first abnormal frame } 
                \State $Collaborators \gets$ first valid subset from recursive split of $Group_{1,2}$ 
                \State Update $\mathbf{c}_{\text{normal}}$, $N$, and $\mathbf{c}_{\text{abnormal}}$
            \EndIf
            \State $D_1^i \gets \operatorname{Perception}(Collaborators)$ \State $D_{\text{ref}} \gets D_1^i$
            \State $D_1^i$ as \textit{PerceptionOutput}
        \EndIf

    \Else

        \State Collaborative perception excluding $Attackers$
    \EndIf

\EndWhile
\end{algorithmic}
\end{algorithm}

\section{The Pseudocode of PRBI Algorithm}
\label{appendix:prbi}

In this section, we provide a detailed pseudocode of the proposed Pseudo-Random Bayesian Inference (PRBI) algorithm.  
The PRBI framework is designed to detect malicious vehicles in collaborative perception with minimal verification attempts, thereby safeguarding the victim vehicle's perception performance under adversarial conditions.  

The full algorithm is outlined in~\cref{alg:prbi}, which formalizes the PRBI defense mechanism for adversarial collaborative perception.

\section{Theoretical Proof}
\label{appendix:proof}
\subsection{Proof of Theorem 1}
\label{appendix:numproof}
In \textbf{Soft Sampling}, we adopt a pseudo-random grouping strategy that assigns the top $\left\lfloor m \right\rfloor$ most suspicious vehicles into one group, and the remaining vehicles into the second group. A key question naturally arises --- \textit{\textbf{Can $m$ converge to the true number of attackers $k$ ?}} This subsection presents a proof showing why pseudo-random grouping ensures that $m$ will indeed converge to the true attacker count $k$. 

Assume the estimated number of attackers at frame $i$ is $m$. We aim to prove that the estimate at the next frame, $m^\prime$, will move toward $k$. The proof is divided into two parts:\\
\textbf{(1) When $m<k$, then $m^\prime$ will increase:} According to the definition of soft sampling, at frame $i+1$, the grouping consists of ``$m$ suspicious vehicles + $(n-m)$ remaining vehicles". Since $m<k$, there must still be at least one attacker in the remaining group. As a result, the normal detection count $\beta$ for both groups will not increase and remains unchanged. The new estimate $m^\prime$ is:
\begin{equation}
    m^\prime = \log_2\left(\frac{n \cdot (N+1)}{\sum \beta} \right) > \log_2\left(\frac{n \cdot N}{\sum \beta} \right) = m.
    \label{eq:proof_1}
\end{equation}
Thus, part (1) is proven based on~\cref{eq:proof_1}.\\
\textbf{(2) When $m \geq k$, then $m^\prime$ will decrease:} Again, the frame $i+1$ grouping consists of ``$m$ suspicious vehicles + $(n-m)$ remaining vehicles". However, this time, the remaining group contains no attackers. The normal detection count for the first group remains unchanged, while the second group contributes $n-m$ additional normal detections. The updated estimate becomes:
\begin{equation}
    m^\prime=\log_2\left(\frac{n \cdot (N+1)}{n-m+\sum\beta} \right).
    \label{eq:m'}
\end{equation}
To prove part (2), we must prove the next inequality:
\begin{equation}
    m^\prime=\log_2\left(\frac{n \cdot (N+1)}{n-m+\sum\beta} \right)<\log_2\left(\frac{n \cdot (N)}{\sum\beta} \right)=m.
    \label{eq:origin(2)}
\end{equation}
Removing the logarithm and rewriting the right-hand side using $2^m$, the~\cref{eq:origin(2)} becomes:
\begin{equation}
    \frac{n \cdot (N+1)}{n-m+\sum\beta}<2^m.
\end{equation}
Divide by $\sum\beta$ to get:
\begin{equation}
\frac{2^m+n/\sum\beta}{1+(n-m)/\sum\beta}<2^m.
\end{equation}
Simplifying further and rearranging terms, the goal is to prove:
\begin{equation}
m+\frac{m}{2^m-1}<n.
\label{eq:keyinequa}
\end{equation}
Define $f(m)=m+\frac{m}{2^m-1}$ for $m \in [1, n-1]$. Taking the derivative of $f(m)$ and we can easily know that $f(m)$ is increasing for $m \in [1, n-1]$. Thus, we have:
\begin{equation}
f(m)\leq f(n-1)=n+\frac{2n-2^n}{2^n-2}.
\end{equation}
When $n=2$, $m$ can only be 1. In this special case, equality holds, and $m^\prime$ always stays at 1. When $n>2$, since $2n-2^n<0$, we have:
\begin{equation}
f(m) \leq  n+\frac{2n-2^n}{2^n-2} <n.
\label{eq:last_eq}
\end{equation}
Based on~\cref{eq:last_eq}, the~\cref{eq:keyinequa} holds and part (2) is proven.

\subsection{Proof of Theorem 2}
\label{appendix:proofodround}
In~\cref{appendix:numproof}, we have proven that under the pseudo-random grouping strategy, the estimated number of attackers $m$ will always converge toward the true number of attackers $k$. In practical applications, it is essential to address another important issue --- \textit{\textbf{How should $m$ be rounded ?}} Since $m$ tends to converge to $k$, directly applying standard rounding (i.e., rounding to the nearest integer) when evaluating suspicious malicious vehicles may suffice. However, during the \textbf{pseudo-random grouping} process, $m$ may not yet have fully converged, and different rounding strategies can influence the convergence outcome. We now provide a detailed derivation of these effects under the same settings and leveraging the proven proof \textbf{(1), (2)} in~\cref{appendix:numproof}:

\textbf{(a) Rounding to the nearest integer $\left\lfloor m \right\rceil$:} When $m \in (-\infty, k - 0.5)$, we have $\left\lfloor m \right\rceil < k$. According to proof \textbf{(1)}, the next frame's $m'$ will increase.
When $m \in [k - 0.5, +\infty)$, $\left\lfloor m \right\rceil \geq k$, and $m'$ will decrease based on proof \textbf{(2)}.
Therefore, $m$ ultimately converges to approximately $k - 0.5$. The subsequent derivations are carried out in the same way.

\textbf{(b) Ceiling $\left\lceil m \right\rceil$:} When $m \in (-\infty, k - 1)$, we have $\left\lceil m \right\rceil < k$, and $m'$ will increase.
When $m \in [k - 1, +\infty)$, $\left\lceil m \right\rceil \geq k$, and $m'$ will decrease.
Thus, $m$ ultimately converges to approximately $k - 1$. Note that when the true number of attackers $k = 1$, $m$ will not tend toward 0, because the argument of a logarithm in~\cref{eq:proof_1} can never be one. In this case, apart from the malicious vehicle, each of the remaining $(n - 1)$ benign vehicles undergoes one successful normal detection per frame. Thus, the final $m$ is:
\begin{equation}
    m  \cong \log_2\left(\frac{n\cdot N}{(n-1)\cdot N} \right) = \log_2(\frac{n}{n-1}).
\end{equation}

\textbf{(c) Floor $\left\lfloor m \right\rfloor$:} When $m \in (-\infty, k)$, we have $\left\lfloor m \right\rfloor < k$, and $m'$ will increase.
When $m \in [k, +\infty)$, $\left\lfloor m \right\rfloor \geq k$, and $m'$ will decrease.
Hence, $m$ ultimately converges to $k$.

In conclusion, \textbf{only by applying floor rounding ($\left\lfloor m \right\rfloor$) in the pseudo-random grouping strategy can we ensure that $m$ converges to $k$}.

\section{Experimental Results}
\subsection{Experimental Details}
\label{appendix:expdetails}
\textbf{Adversarial Optimization.} Perturbations are added to all malicious vehicles’ features and optimized jointly. For adversarial attacks, the step size is set to 0.1, the perturbation stealth metric to 0.3, and the number of iterations to 15.

\textbf{Baselines and Implementation.} We randomly select 5 scenarios from the V2X-Sim test set, each consisting of 100 frames. Collaborative perception scenarios involve 6 collaborators (1 RSU and 5 vehicles, with the first vehicle being the ego/victim). Detection model training follows the default setup in DiscoNet~\cite{li2021learning}. For baseline defenses, ROBOSAC~\cite{li2023among} strictly follows the official configuration and always selects the maximum possible number of collaborators for joint perception to ensure fairness and stability. PASAC~\cite{hu2025cp} also adopts its official hyperparameter settings for the consistency loss module. All baseline implementations are reproduced under the same perception backbone, communication setup, and attack conditions for a fair comparison.

\textbf{Defense Hyperparameters.} The Jaccard similarity threshold $\epsilon$ is 0.35, the verification window size $w$ is 10, the confidence level $\alpha$ for the T-test is 0.01, and the probability weights $\gamma$ and $\lambda$ are set to 0.35 and 0.65, respectively. The setting of the similarity threshold is based on the following considerations: (1) a relatively large threshold may occasionally misclassify benign vehicles as malicious in a single frame, but such errors can be corrected in subsequent frames; (2) setting the threshold too small risks misclassifying malicious vehicles as benign, yet this is tolerable in our framework. Specifically, a false negative occurs when malicious vehicles exhibit high similarity, implying that the attack has not significantly impacted perception and thus does not cause irreversible consequences. Conversely, effective attacks induce a sustained drop in similarity, and our probabilistic updating mechanism ensures that such malicious vehicles will eventually be identified. Combined with the experimental results in~\cref{appendix:hypothesisverification} (where adversarial cases rarely exceed 0.3), we adopt 0.35 as a balanced threshold.

\begin{table}[!ht]\fontsize{8.5pt}{10pt}\selectfont
    \centering
    \caption{Comprehensive experimental results of different test parameters on defense effect. \textit{ID Rate}: malicious vehicle identification rate. \textit{MC Rate}: benign vehicle misclassification rate.}
    \label{tab:wholepara}
    \setlength{\tabcolsep}{3pt}
    \begin{tabular}{@{}l|cc|cc|c|c|cc@{}}
        \toprule
        \multirow{2}{*}{\textbf{Settings}} & \multicolumn{2}{c|}{\textbf{Count}} & \multicolumn{2}{c|}{\textbf{Avg. Count}} & \multirow{2}{*}{\textbf{Avg.}} & \textbf{Avg.} & \textbf{ID} & \textbf{MC} \\
        \cmidrule(lr){2-3} \cmidrule(lr){4-5}
        & \textbf{Min} & \textbf{Max} & \textbf{Min} & \textbf{Max} & & \textbf{Frames} & \textbf{Rate} & \textbf{Rate} \\
        \midrule
        \multicolumn{9}{c}{\textbf{\emph{Confidence Level (20\% Attackers)}}} \\
        \midrule 
        Conf. 0.20 & 2 & 2 & 2.00 & 2.00 & 2.00 & 2.53 & 100\% & 0\% \\
        Conf. 0.15 & 2 & 2 & 2.00 & 2.00 & 2.00 & 2.27 & 100\% & 0\% \\
        Conf. 0.10 & 2 & 2 & 2.00 & 2.00 & 2.00 & 2.29 & 100\% & 0\% \\
        Conf. 0.05 & 2 & 2 & 2.00 & 2.00 & 2.00 & 2.19 & 100\% & 0\% \\
        Conf. 0.01 & 2 & 2 & 2.00 & 2.00 & 2.00 & 2.25 & 100\% & 0\% \\
        \midrule
        \multicolumn{9}{c}{\textbf{\emph{Confidence Level (40\% Attackers)}}} \\
        \midrule 
        Conf. 0.20 & 2 & 4 & 2.00 & 3.02 & 2.38 & 3.13 & 100\% & 9\% \\
        Conf. 0.15 & 2 & 4 & 2.00 & 3.00 & 2.37 & 3.04 & 100\% & 8\% \\
        Conf. 0.10 & 2 & 4 & 2.00 & 2.96 & 2.35 & 2.98 & 100\% & 8\% \\
        Conf. 0.05 & 2 & 4 & 2.00 & 2.91 & 2.34 & 2.77 & 100\% & 4\% \\
        Conf. 0.01 & 2 & 4 & 2.00 & 2.88 & 2.35 & 2.77 & 100\% & 6\% \\
        \midrule
        \multicolumn{9}{c}{\textbf{\emph{Confidence Level (60\% Attackers)}}} \\
        \midrule 
        Conf. 0.20 & 2 & 6 & 2.00 & 4.22 & 2.61 & 3.99 & 100\% & 0\% \\
        Conf. 0.15 & 2 & 6 & 2.00 & 4.12 & 2.62 & 3.51 & 100\% & 0\% \\
        Conf. 0.10 & 2 & 6 & 2.00 & 3.94 & 2.53 & 3.44 & 100\% & 0\% \\
        Conf. 0.05 & 2 & 6 & 2.00 & 4.26 & 2.67 & 3.40 & 100\% & 0\% \\
        Conf. 0.01 & 2 & 6 & 2.00 & 4.06 & 2.61 & 3.36 & 100\% & 0\% \\
        \midrule
        \multicolumn{9}{c}{\textbf{\emph{Confidence Level (80\% Attackers)}}} \\
        \midrule 
        Conf. 0.20 & 2 & 8 & 2.00 & 6.06 & 2.50 & 10.01 & 100\% & 0\% \\
        Conf. 0.15 & 2 & 8 & 2.00 & 5.86 & 2.51 & 9.31 & 100\% & 0\% \\
        Conf. 0.10 & 2 & 8 & 2.00 & 6.24 & 2.80 & 6.58 & 100\% & 0\% \\
        Conf. 0.05 & 2 & 8 & 2.00 & 6.04 & 2.77 & 6.51 & 100\% & 0\% \\
        Conf. 0.01 & 2 & 8 & 2.00 & 5.70 & 2.86 & 4.27 & 100\% & 0\% \\
        \midrule
        \multicolumn{9}{c}{\textbf{\emph{Window Size (20\% Attackers)}}} \\
        \midrule 
        Size 10 & 2 & 2 & 2.00 & 2.00 & 2.00 & 2.25 & 100\% & 0\% \\
        Size 8 & 2 & 2 & 2.00 & 2.00 & 2.00 & 2.37 & 100\% & 0\% \\
        Size 6 & 2 & 2 & 2.00 & 2.00 & 2.00 & 2.35 & 100\% & 0\% \\
        Size 4 & 2 & 2 & 2.00 & 2.00 & 2.00 & 2.33 & 100\% & 0\% \\
        \midrule
        \multicolumn{9}{c}{\textbf{\emph{Window Size (40\% Attackers)}}} \\
        \midrule 
        Size 10 & 2 & 4 & 2.00 & 2.88 & 2.35 & 2.77 & 100\% & 6\% \\
        Size 8 & 2 & 4 & 2.00 & 3.12 & 2.46 & 2.80 & 100\% & 7\% \\
        Size 6 & 2 & 4 & 2.00 & 3.00 & 2.39 & 2.84 & 100\% & 4\% \\
        Size 4 & 2 & 4 & 2.00 & 3.02 & 2.40 & 2.94 & 100\% & 11\% \\
        \midrule
        \multicolumn{9}{c}{\textbf{\emph{Window Size (60\% Attackers)}}} \\
        \midrule 
        Size 10 & 2 & 6 & 2.00 & 4.06 & 2.61 & 3.36 & 100\% & 0\% \\
        Size 8 & 2 & 6 & 2.00 & 4.08 & 2.62 & 3.33 & 100\% & 0\% \\
        Size 6 & 2 & 6 & 2.00 & 4.32 & 2.69 & 3.34 & 100\% & 0\% \\
        Size 4 & 2 & 6 & 2.00 & 4.14 & 2.63 & 3.41 & 100\% & 0\% \\
        \midrule
        \multicolumn{9}{c}{\textbf{\emph{Window Size (80\% Attackers)}}} \\
        \midrule 
        Size 10 & 2 & 8 & 2.00 & 5.70 & 2.86 & 4.27 & 100\% & 0\% \\
        Size 8 & 2 & 8 & 2.00 & 6.16 & 2.91 & 4.60 & 100\% & 0\% \\
        Size 6 & 2 & 8 & 2.00 & 6.10 & 2.89 & 4.57 & 100\% & 0\% \\
        Size 4 & 2 & 8 & 2.00 & 6.34 & 2.88 & 4.89 & 100\% & 0\% \\
        \bottomrule
    \end{tabular}
    \vspace{-3mm}
\end{table}

\subsection{Varying Test Parameters}
\label{sec:varyingtestpara}
In the scenario where the total number of vehicles is set to 5, the results of the hypothesis testing parameter ($\alpha$ and $w$) experiments for all attacker ratios are presented in~\cref{tab:wholepara}. We document the total number of frames required to achieve convergence under each setting \textit{\textbf{(Average Converge Frames)}}, the identification rate of malicious vehicles upon convergence \textit{\textbf{(Identification Rate)}}, and the misclassification rate of benign vehicles upon convergence \textit{\textbf{(Misclassification Rate)}}. The results show that regardless of the proportion of attackers, the convergence speed is always inversely proportional to the confidence level $\alpha$ and directly proportional to the window size $w$. For instance, when the $\alpha$ is set to 0.2, the average maximum number of frames required for convergence is 10.01 frames; while the $\alpha$ is reduced to 0.01, the average minimum number of frames needed for convergence is only 2.25 frames. Additionally, the experimental results show that under all test conditions, the identification rate of malicious vehicles reaches 100\%, fully verifying the high robustness of the proposed method.

\subsection{Robustness to Intermittent Attacks}
\label{sec:intermittent}

We further evaluate PRBI under intermittent attacks, where adversarial perturbations are injected periodically every 1, 3, or 5 frames.
The total number of vehicles is set to $n=5$, and all other experimental settings remain consistent with the main text.

PRBI is inherently robust to intermittent attacks due to its temporal Bayesian updating mechanism.
Even when attacks occur sporadically, the posterior benign probability of malicious vehicles decreases cumulatively over time, leading to their progressive exclusion from the collaborator subset. As shown in~\cref{tab:intermittent}, PRBI consistently achieves 100\% attacker identification with 0\% benign misclassification under all attack frequencies.
Detection performance remains stable across both V2VNet and DiscoNet backbones, demonstrating strong robustness against temporally sparse adversarial behavior.

\begin{table}[!t]\small
\centering
\caption{Defense performance under intermittent attacks.}
\label{tab:intermittent}
\setlength{\tabcolsep}{2pt}
\begin{tabular}{l|c|c|c|c}
\hline
 & \multicolumn{2}{c|}{V2VNet} & \multicolumn{2}{c}{DiscoNet} \\
Methods & AP@0.5 / 0.7 & ID / MC & AP@0.5 / 0.7 & ID / MC \\
\hline
1 frame  & 68.91 / 65.57 & 100\% / 0\% & 69.37 / 65.89 & 100\% / 0\% \\
3 frames & 70.97 / 65.91 & 100\% / 0\% & 69.12 / 66.27 & 100\% / 0\% \\
5 frames & 73.52 / 67.04 & 100\% / 0\% & 70.43 / 66.91 & 100\% / 0\% \\
\hline
\end{tabular}
\end{table}

\subsection{Sensitivity Analysis of Similarity Threshold $\epsilon$}
\label{sec:eps_analysis_clean}

We evaluate the sensitivity of PRBI to the similarity threshold $\epsilon$ under standard adversarial settings. The total number of vehicles is set to $n=5$, and other experimental configurations remain consistent with the main text.

As shown in~\cref{tab:eps_ablation_clean}, PRBI maintains 100\% attacker identification and 0\% benign misclassification across a wide range of thresholds $\epsilon \in [0.30, 0.50]$.
This indicates that the correctness of PRBI does not critically depend on precise threshold tuning. From a performance perspective, smaller $\epsilon$ values lead to slightly faster convergence but may introduce marginal reductions in detection precision, whereas larger $\epsilon$ values enforce stricter consistency validation, resulting in improved stability at the cost of modestly increased convergence latency. This behavior reflects a controllable strictness--efficiency trade-off.

Overall, the robustness of PRBI primarily stems from its probabilistic temporal accumulation mechanism rather than the specific value of $\epsilon$.
Empirically, $\epsilon=0.35$ achieves the best balance between detection accuracy and convergence speed.

\begin{table}[!ht]\small
\centering
\caption{Sensitivity of PRBI to similarity threshold $\epsilon$ under standard adversarial settings.}
\label{tab:eps_ablation_clean}
\setlength{\tabcolsep}{4pt}
\begin{tabular}{c|c|c}
\hline
$\epsilon$ & AP@0.5 / 0.7 & Avg. Convergence Frames \\
\hline
0.30 & 69.17 / 64.14 & 2.41 \\
0.35 & 69.44 / 66.00 & 2.74 \\
0.40 & 68.61 / 64.14 & 2.90 \\
0.45 & 69.40 / 65.80 & 2.93 \\
0.50 & 67.47 / 64.96 & 2.88 \\
\hline
\multicolumn{3}{l}{\footnotesize ID Rate: 100\%, MC Rate: 0\% across all settings} \\
\hline
\end{tabular}

\end{table}